\pdfoutput=1

\documentclass[11pt]{article}

\usepackage[final]{acl}

\usepackage{times}
\usepackage{latexsym}

\usepackage[T1]{fontenc}
\usepackage[utf8]{inputenc}

\usepackage{microtype}
\usepackage{inconsolata}

\usepackage{graphicx}
\usepackage{amsmath}
\usepackage{amssymb}  
\usepackage{amsfonts}
\usepackage{algorithmic}
\usepackage{algorithm}
\usepackage{booktabs}  
\usepackage{multirow}
\usepackage{xspace}

\usepackage{pifont}
\usepackage{multirow}
\usepackage{array}
\usepackage{booktabs}
\usepackage{colortbl}
\usepackage{color}
\usepackage{enumitem}
\usepackage{tcolorbox}    
\usepackage[frozencache,cachedir=minted-cache]{minted}
\usepackage{xcolor}       
\tcbuselibrary{breakable} 
\usepackage{ulem}
\usepackage[capitalize,noabbrev]{cleveref}

\title{Self-Supervised Prompt Optimization}

\author{%
 \textbf{Jinyu Xiang}$^{1}\footnotemark[1]$,
 \textbf{Jiayi Zhang}$^{1,2} \thanks{These authors contributed equally to this work.}$,  
 \textbf{Zhaoyang Yu}$^{1}$,
 \textbf{Xinbing Liang}$^{1}$,
 \textbf{Fengwei Teng}$^{2}$, \\
 \textbf{Jinhao Tu}$^{3}$, 
 \textbf{Fashen Ren}$^{2}$,
 \textbf{Xiangru Tang}$^{4}$,
 \textbf{Sirui Hong}$^{1}$,
 \textbf{Chenglin Wu}$^{1} \footnotemark[2]$,
 \textbf{Yuyu Luo}$^{2} \thanks{Corresponding authors: Chenglin Wu (E-mail: alexanderwu@deepwisdom.ai), Yuyu Luo (E-mail:yuyuluo@hkust-gz.edu.cn)}$ ,
   \vspace{.5em} 
  \\
  $^1$DeepWisdom, 
  $^2$The Hong Kong University of Science and Technology (Guangzhou), \\
  $^3$Independent Researcher,
  $^4$Yale University,
}

\usepackage{xspace}

\usepackage{amsmath}
\usepackage{amssymb}

\newcommand{\eg}{{\em e.g.,}\xspace}

\newcommand{\ours}{\textbf{SPO}\xspace}
\newcommand{\argmax}{\mathop{\arg\max}}

\definecolor{bleudefrance}{rgb}{0.19, 0.55, 0.91}

\usepackage{xcolor}

\definecolor{prompt}{RGB}{241,204,185}
\definecolor{metaprompt}{RGB}{218,218,218}
\definecolor{output}{RGB}{185,240,237}
\definecolor{response}{RGB}{185,240,205}

\newcommand{\highlight}[2]{%
    \raisebox{0pt}[0pt][0pt]{\colorbox{#1}{#2}}%
}
\begin{document}
\maketitle

\begin{abstract}
Well-designed prompts are crucial for enhancing Large language models' (LLMs) reasoning capabilities while aligning their outputs with task requirements across diverse domains. However, manually designed prompts require expertise and iterative experimentation. 
While existing prompt optimization methods aim to automate this process, they rely heavily on external references such as ground truth or by humans, limiting their applicability in real-world scenarios where such data is unavailable or costly to obtain. 
To address this, we propose \textbf{S}elf-Supervised \textbf{P}rompt \textbf{O}ptimization (\ours), a cost-efficient framework that discovers effective prompts for both closed and open-ended tasks without requiring external reference.
Motivated by the observations that prompt quality manifests directly in LLM outputs and LLMs can effectively assess adherence to task requirements, we derive evaluation and optimization signals purely from output comparisons.
Specifically, \ours selects superior prompts through pairwise output comparisons evaluated by an LLM evaluator, followed by an LLM optimizer that aligns outputs with task requirements.
Extensive experiments demonstrate that \ours outperforms state-of-the-art prompt optimization methods, achieving comparable or superior results with significantly lower costs (e.g., \textbf{1.1\% to 5.6\%} of existing methods) and fewer samples (e.g., three samples). The code is available at \href{https://github.com/FoundationAgents/SPO}{https://github.com/FoundationAgents/SPO}.
\end{abstract}
\section{Introduction}

\begin{figure}[t!]
	\centering
\includegraphics[width=\linewidth]{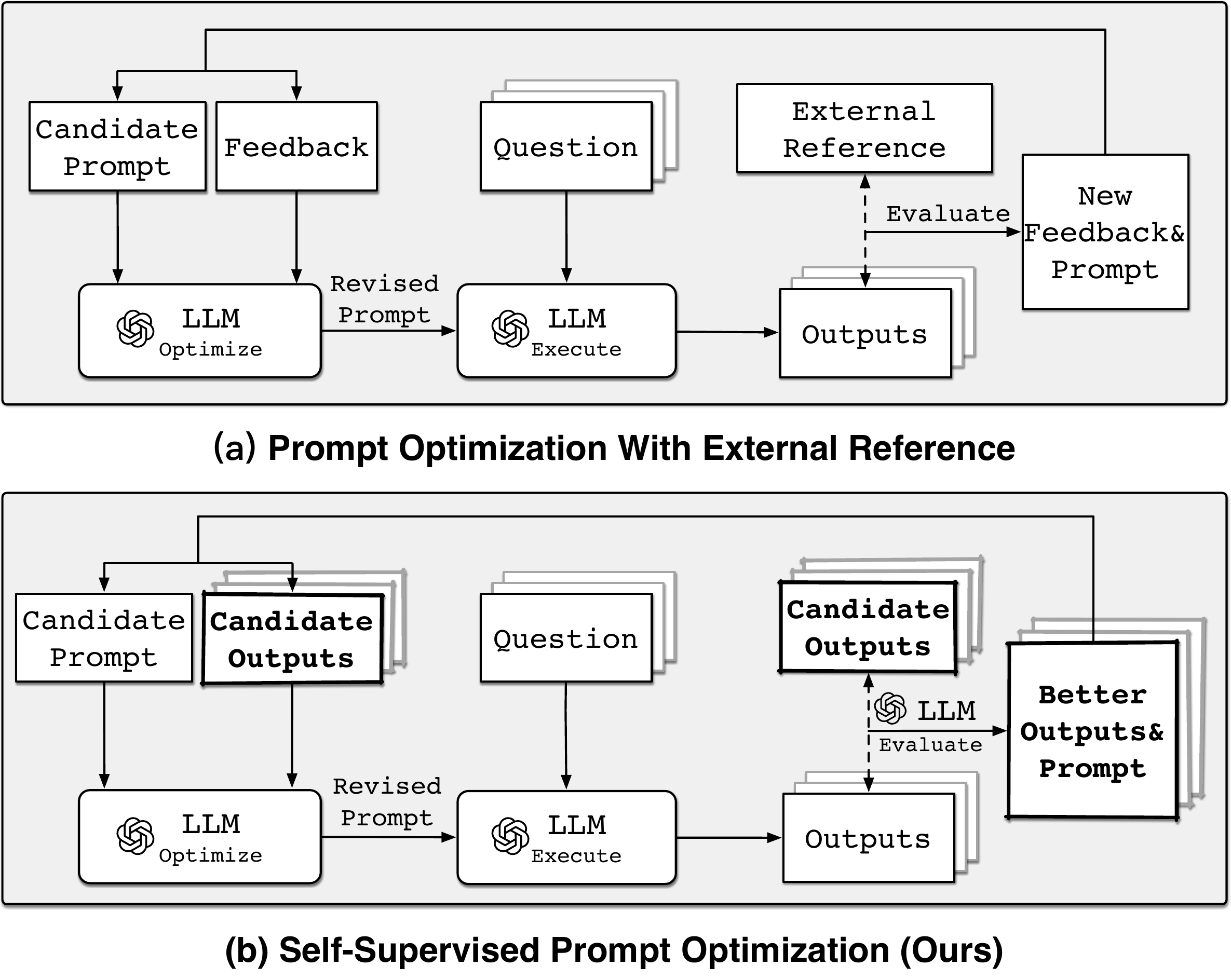}
        \vspace{-1em}
        \caption{\textbf{Comparison of Prompt Optimization Methods.} (a) illustrates the traditional prompt optimization process with external reference, where feedback from the ground truth of humans is used to iteratively improve the best prompt. (b) presents our proposed self-supervised prompt optimization, which utilizes pairwise comparisons of LLM's own outputs to optimize prompts without relying on external reference.}
	\vspace{-1em}
        \label{fig:contrast}
\end{figure}

\begin{figure*}[t!]
	\centering
	\includegraphics[width=\linewidth]{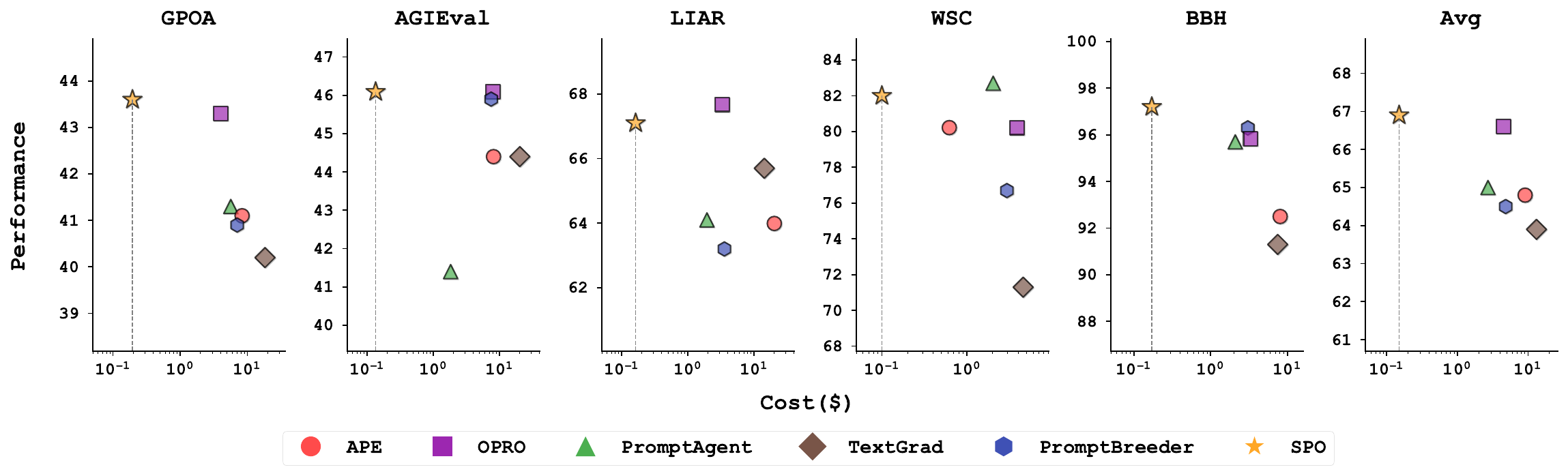}
        \vspace{-1em}
	\caption{\textbf{Comparison of Performance ($y$-axis) and Optimization Costs in Dollars ($x$-axis) across Six Prompt Optimization Methods.}
     \ours demonstrates competitive performance, consistently ranking among the top two methods while maintaining significantly lower costs (ranging from 1.1\% to 5.6\% of the costs incurred by other methods) across all datasets.
     }
    \label{fig:performance-cost}

\end{figure*}

As large language models (LLMs) continue to advance, well-designed prompts have become critical for maximizing their reasoning capabilities~\cite{wei2022COT, hua2024step, deng2023rephrase} and ensuring alignment with diverse task requirements.
However, creating effective prompts often requires substantial trial-and-error experimentation and deep task-specific knowledge.

To address this challenge, researchers have explored Prompt Optimization (PO) methods that use LLMs' own capabilities to automatically improve prompts. PO advances beyond traditional prompt engineering, by providing a more systematic and efficient approach to prompt design. As shown in Figure~\ref{fig:contrast}(a), these methods typically involve an iterative process of prompt optimization, execution, and evaluation. The design choices for these components significantly influence optimization effectiveness and efficiency.
Existing approaches have demonstrated success with both numerical evaluation mechanisms~\cite{xin2024pa, yang2023opro, chris2024pb} and textual ``gradient'' optimization strategies~\cite{mert2024textgrad}. Through these innovations, PO methods have shown promise in reducing manual effort while enhancing task performance~\cite{ zhang2024aflow, zhou2024zepo, he2025crispo}.

Despite their potential, existing PO methods face significant challenges in real-world scenarios, as discussed below. First, current methods often depend heavily on external references for evaluation. Methods using ground truth for evaluation~\cite{yang2023opro, chris2024pb, mert2024textgrad,reid2023protegi} require large amounts of annotated data to assess prompt quality, yet such standard answers are often unavailable in many practical applications, especially for open-ended tasks. Similarly, methods relying on human~\cite{yong2024promst, lin2024apohf} require manual evaluations or human-designed rules to generate feedback, which is time-consuming and contradicts the goal of automation.
Second, existing methods typically require evaluating prompts on numerous samples to obtain reliable feedback, leading to substantial computational overhead~\cite{xin2024pa, chris2024pb}.

At the core of these challenges lies the absence of reliable and efficient reference-free methods for assessing prompt quality. Analysis of LLM behavior reveals two key insights that inform our approach. First, prompt quality inherently manifests in model outputs, as evidenced by how different prompting strategies significantly influence both reasoning paths~\cite{wei2022COT, deng2023rephrase} and response features~\cite{lei2024character, schmidgall2025agentlaboratory}. Second, extensive studies on LLM-as-a-judge have demonstrated their effectiveness in evaluating output adherence to task requirements~\cite{lianmin2023mtbench, dawei2024laajsurvey}. These observations suggest that by leveraging LLMs' inherent ability to assess outputs that naturally reflect prompt quality, reference-free prompt optimization becomes feasible.

Motivated by these insights, we propose a cost-efficient framework that generates evaluation and optimization signals purely from LLM outputs, similar to how self-supervised learning derives training signals from data. We term this approach \textbf{S}elf-Supervised \textbf{P}rompt \textbf{O}ptimization (\ours). As shown in Figure \ref{fig:contrast}, \ours builds upon the fundamental Optimize-Execute-Evaluate loop while introducing several innovative mechanisms: 

(1) \textbf{\textit{Output as Pairwise Evaluation Reference}}: At its core, \ours employs a pairwise comparison approach that assesses the relative quality of outputs from different prompts. This evaluation mechanism leverages LLM's inherent capability to understand task requirements, validating optimization effectiveness without external references.

(2) \textbf{\textit{Output as Optimization Guidance}}: \ours optimizes prompts through LLM's understanding of better solutions for the current best output. Rather than relying on explicit optimization signals, this process naturally aligns prompt modifications with the model's comprehension of optimal task solutions.

\textbf{Contributions.}
Our main contributions are as follows:

(1) \textbf{Self-Supervised Prompt Optimization Framework.} We introduce \ours, a novel framework that leverages pairwise comparisons of LLM's outputs to guide prompt optimization without requiring external reference.

(2) \textbf{Cost-effective Optimization.} 
\ours optimizes prompts with minimal computational overhead (\$0.15 per dataset) and sample requirements (3 samples), significantly reducing resource demands.

(3) \textbf{Extensive Evaluation.} As shown in Figure~\ref{fig:performance-cost}, \ours requires only \textbf{1.1\% to 5.6\%} of the cost of state-of-the-art methods while maintaining superior performance across both closed and open-ended tasks.

\section{Preliminary}
\label{sec:pre}

\subsection{Problem Definition} 
\label{sec:problem-formulation}

\textbf{Prompt Optimization} aims to automatically enhance the effectiveness of a prompt for a given task type $t\in \mathcal{T}$. For a given type $t$, we observe a dataset $D_t=\{(Q_i, G_i)\}_{i=1}^{N_t}$, which each pair consists of an input query $Q_i$ and the optional ground truth $G_i$. A candidate prompt for type $t$ is denoted $P_t\in \mathcal{P}$, and the goal is to find a optimal prompt.

\begin{equation}
    P_t^*=\argmax_{P_t \in \mathcal{P}} \mathbb{E}_{(Q,G)\sim D_t}\Bigl[\phi_{\text{eval}}\!\bigl(\phi_{\text{exe}}(Q, P_t), G\bigr)\Bigr].
\end{equation}

As illustrated in Figure~\ref{fig:contrast}, this optimization process typically involves three fundamental functions:
(1) Optimization function ($\phi_{\text{opt}}$): generates a revised prompt based on the candidate prompt; 
(2) Execution function ($\phi_{\text{exe}}$): applies the revised prompt with an LLM to produce output $O$, consisting of a reasoning path and a final answer; 
(3) Evaluation function ($\phi_{\text{eval}}$):  assesses the quality of $O$ and provides feedback $F$ to guide further optimization, refining the candidate prompts iteratively.

Among these functions, the evaluation function plays a pivotal role as its output (feedback $F$) guides the assessment and improvement of prompts.
We will discuss the evaluation framework for prompt optimization in Section~\ref{sec:feedback}.

\subsection{Evaluation Framework in  Prompt Optimization}
\label{sec:feedback}

\begin{figure}[t!]
	\centering
	\includegraphics[width=\linewidth]{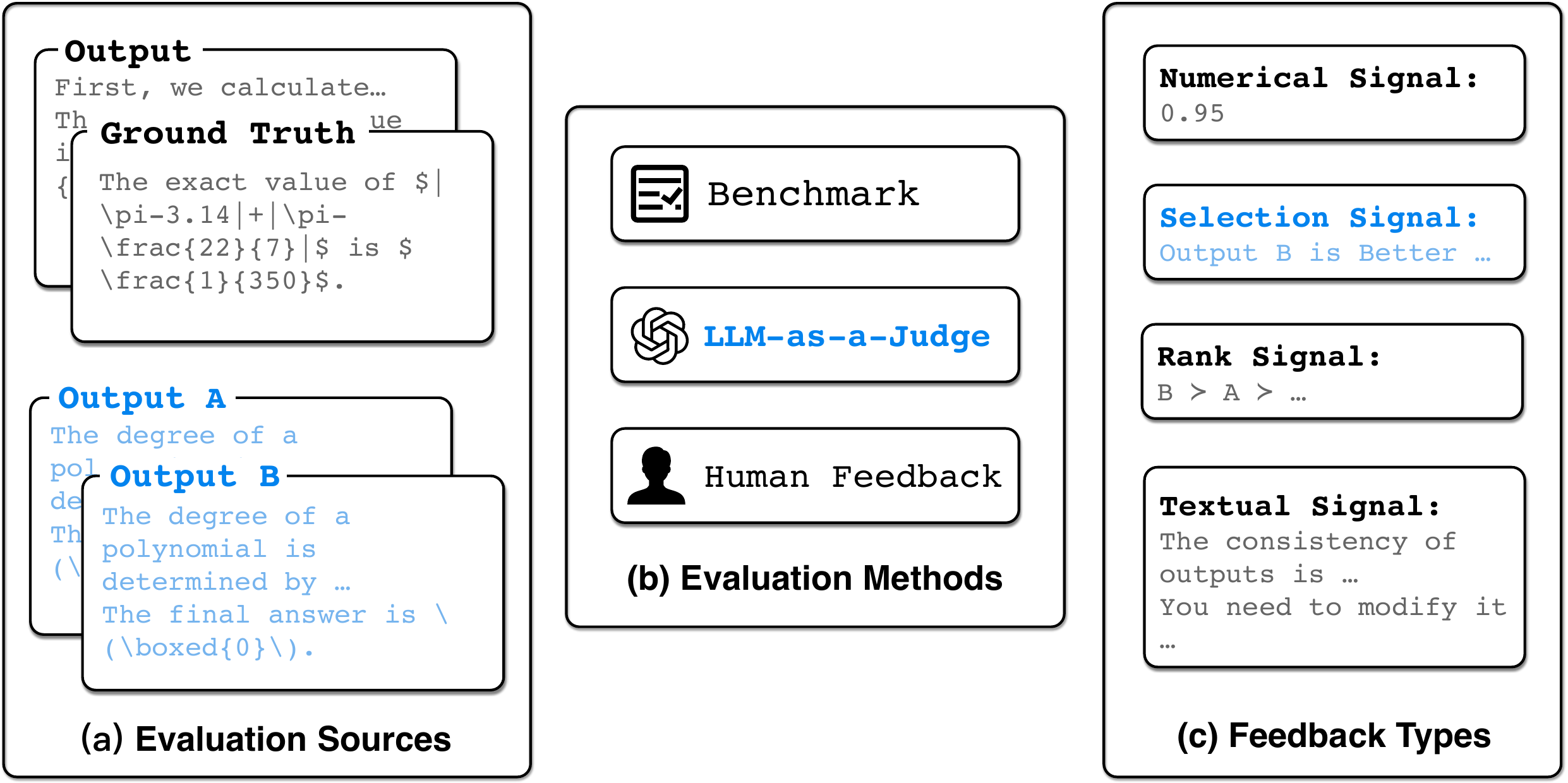}
        \vspace{-2em}
	\caption{Components of the Evaluation Framework for Prompt Optimization. (a) Evaluation Sources: Compares different outputs, including ground truth and model-generated outputs, to assess quality.
(b) Evaluation Methods: Showcases various evaluation techniques, including benchmark comparisons, LLM-as-a-Judge, and human feedback.
(c) Feedback Types: Showcases a range of feedback. For clarity, the rank-signal example now compares only Output A and Output B.
The \textit{\textcolor{bleudefrance}{\textbf{blue}}} in (a), (b), and (c) indicate the specific evaluation approach selected for \ours.
}
	\label{fig:components}
\end{figure}

This section outlines our evaluation framework for prompt optimization, covering three key components: evaluation sources, evaluation methods, and feedback types, as shown in Figure~\ref{fig:components}. We conclude by introducing our selected evaluation framework for \ours.

\textbf{Evaluation Sources}
As shown in Figure~\ref{fig:components}(a), two primary sources can be used for evaluation: LLM-generated outputs and task-specific ground truth. These sources provide the basis for assessing prompt performance.

\textbf{Evaluation Methods}
The evaluation method defines how the evaluation sources are assessed and the associated costs. Three common methods are used:
(1) \textit{Benchmark} relies on predefined metrics~\cite{mirac2023bbh, david2023gpqa} or rules~\cite{yong2024promst}. 
(2) \textit{LLM-as-a-judge}~\cite{lianmin2023mtbench} leverage LLMs capability to understand and assess outputs based on task requirements. 
(3) \textit{Human Feedback}~\cite{lin2024apohf} provides the most comprehensive evaluation through direct human assessment of outputs.

While Human Feedback offers the most thorough evaluation by capturing human preferences and task-specific needs, it incurs substantially higher costs than Benchmark or LLM-as-a-judge methods, creating a trade-off between evaluation quality and feasibility.

\textbf{Feedback Types}
Feedback produced by evaluation methods typically take three forms:
(1) \textit{Numerical Feedback} provides quantitative performance measures across the dataset. However, it requires substantial samples for stable evaluation and may overlook instance-specific details~\cite{zhang2024aflow}. 
(2) \textit{Textual Feedback} offers rich, instance-specific guidance through analysis and suggestions, directly generating optimization signals~\cite{mert2024textgrad}.
(3) \textit{Ranking or Selection Feedback} establishes relative quality ordering among outputs through either complete ranking or pairwise comparisons, providing clear optimization direction without requiring absolute quality measures.

\paragraph{Evaluation Framework}
Building on the previous discussion on evaluation's sources, methods, and feedback types, the evaluation framework determines how sources are compared and assessed within the context of prompt optimization. Specifically, we derive two evaluation frameworks to generate feedback $F$ for prompt optimization:

(1) \textbf{Output \textit{vs.} Ground Truth (OvG):} 
    Feedback is generated by comparing the output $O_i$ of query $Q_i$ with ground truth $G_i$:

    \begin{equation}
        f_{\text{OvG}} = \phi_{eval}\Bigl(\phi_{exe}(Q_i, P_t), G_i\Bigr).
    \end{equation}
    
    Although this approach allows for a direct quality assessment through an external reference, it requires well-defined ground truth, making it unsuitable for open-ended tasks where ground truth may not always be available or practical to define.
    
(2) \textbf{Output \textit{vs.} Output (OvO):}
    When ground truth is unavailable, we turn to direct output comparison. The core idea behind OvO is that even in the absence of perfect ground truth, comparing outputs generated by different prompts can offer valuable signals about their relative quality. This method removes the dependency on external references and is particularly useful for open-ended tasks where multiple answers may be valid. It can be formally expressed as:

    \begin{equation}
        f_{\text{OVO}} = \phi_{\text{eval}}\Bigl(\{\phi_{\text{exe}}(Q_i,P_i)\},G_i\Bigr).
    \end{equation}

After introducing the \textbf{OvG} and \textbf{OvO} evaluation frameworks, we emphasize that \textbf{OvO} serves as the core method in Self-Supervised Prompt Optimization (\ours). By comparing outputs generated by different prompts, \textbf{OvO} provides valuable feedback on their relative quality without relying on external reference. This approach aligns with our objective of generating feedback directly from the outputs themselves, thus facilitating iterative optimization in both closed and open-ended tasks.

\section{Self-Supervised Prompt Optimization}
\label{sec:method}

In this section, we first overview our method (Section~\ref{sub:overview}) and then analyze its effectiveness (Section~\ref{sec:method_essence}).

\begin{figure*}[t!]
  \centering
\includegraphics[width=\linewidth]{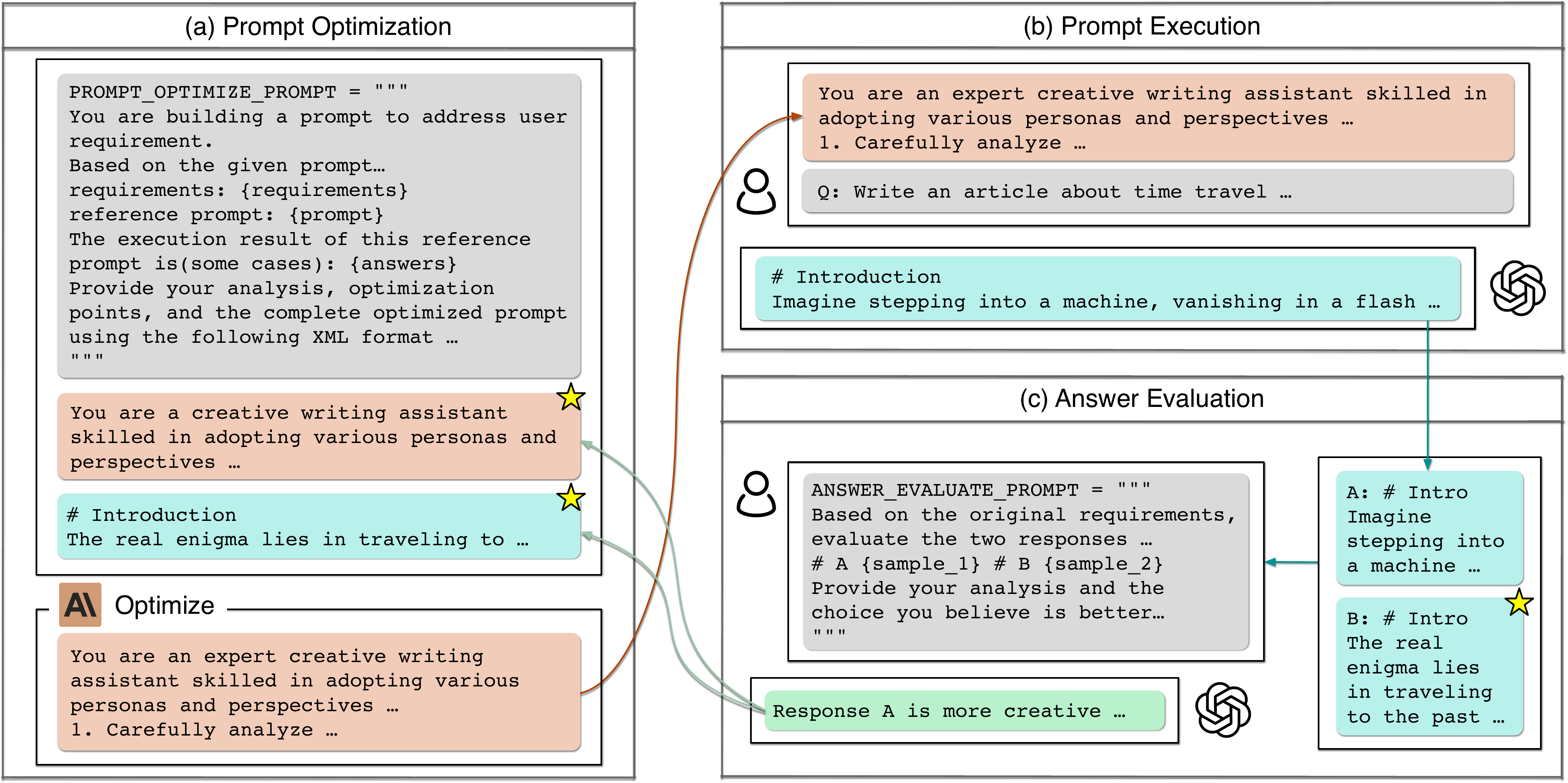}
  \vspace{-2em}
  \caption{A Running Example of \ours Framework: Pairwise evaluation on the \highlight{output}{outputs} selects the \highlight{response}{better} one from corresponding \highlight{prompt}{prompts}. The best \highlight{output}{output} and \highlight{prompt}{prompt} pair are highlited with pentagrams, which will be updated after evaluation. Furthermore, using a case from MT-bench, we show the complete process of \ours's $\phi_{opt}$, $\phi_{exe}$, and $\phi_{eval}$ and corresponding \highlight{metaprompt}{prompt}.}
  \label{fig:main_method}
\end{figure*}

\subsection{An Overview of SPO}
\label{sub:overview}

A core challenge in reference-free prompt optimization is how to construct effective evaluation and optimization signals. We propose Self-Supervised Prompt Optimization (\ours), a simple yet effective framework that retains the basic Optimize-Execute-Evaluate loop while enabling reference-free optimization by leveraging only model outputs as both evaluation sources and optimization guidance.

As shown in Algorithm~\ref{alg:concise-algo-pipo}, \ours operates through three key components and the corresponding prompts are shown in Appendix \ref{appendix:prompt}: 

\begin{itemize}
\item Optimization function ($\phi_{opt}$): Generates new prompts by analyzing the current best prompt and its corresponding outputs.
\item Execution function ($\phi_{exe}$): Applies the generated prompts to obtain outputs.
\item Evaluation function ($\phi_{eval}$): Uses an LLM to compare outputs and determine the superior prompt through pairwise comparisons.
\end{itemize}

This iterative process begins with a basic prompt template (\eg Chain-of-Thought~\cite{wei2022COT}) and a small question set
sampled from the dataset. In each iteration, \ours generates new prompts, executes them, and performs pairwise evaluations of outputs to assess their adherence to task requirements.

The prompt associated with the superior output is selected as the best candidate for the next iteration. The process continues until a predefined maximum number of iterations is reached.

\textbf{A Running Example}
As illustrated in Figure~\ref{fig:main_method}, \ours achieves high efficiency, requiring only 8 LLM calls per iteration with three samples, significantly lower than existing methods~\cite{xin2024pa, chris2024pb, mert2024textgrad, 10720675, yong2023ape}. Despite its simplicity, \ours demonstrates superior performance across a range of tasks, as detailed in Section~\ref{sec:exp}. In the following section, we analyze the theoretical foundations of its effectiveness.

\begin{algorithm}[t!]
\small
\caption{An Overview of \ours.}
\label{alg:concise-algo-pipo}
\begin{algorithmic}[1]
\REQUIRE Dataset $D$
\ENSURE Optimized Prompt $P^*$
\STATE Initialize $P_0$; Sample 3 Questions $Q$ from $D$
\STATE $\text{Best Prompt } P^* \gets P_0$
\STATE $\text{Best Answer } A^* \gets \phi_{exe}(Q, P^*)$
\FOR{$iteration \gets 1$ to $N_{max}$}
    \STATE $P' \gets \phi_{opt}(P^*, A^*)$
    \STATE $A' \gets \phi_{exe}(Q, P')$
    \STATE $optimizationSuccess \gets \phi_{eval}(Q, A', A^*)$
    \IF{$optimizationSuccess$}
        \STATE $P^* \gets P'$
        \STATE $A^* \gets A'$
    \ENDIF
\ENDFOR
\STATE \textbf{return} $P^*$
\end{algorithmic}
\end{algorithm}

\subsection{Understanding the Effectiveness of \ours}
\label{sec:method_essence}

The theoretical foundation of \ours is built upon two key observations:

First, the outputs of LLMs inherently contain rich quality information that directly reflects prompt effectiveness, as evidenced by how step-by-step reasoning paths demonstrate the success of Chain-of-thought prompting~\cite{wei2022COT}. Second, LLMs exhibit human-like task comprehension, enabling them to assess answer quality and identify superior solutions based on task requirements. These complementary capabilities allow SPO to perform prompt evaluation and optimization without external references. These two aspects of utilizing model outputs work together to enable effective prompt optimization:

\textbf{Output as Optimization Guidance}
In terms of $\phi_{opt}$ design, unlike other methods that introduce explicit optimization signals \cite{chris2024pb, mert2024textgrad, reid2023protegi}, $\phi_{opt}$ optimizes directly based on the prompt and its corresponding outputs. The optimization signal stems from the LLMs' inherent ability to assess output quality, while the optimization behavior is guided by its understanding of what constitutes superior solutions. Therefore, even without explicit optimization signals, \ours's optimization essentially guides prompts toward the LLM's optimal understanding of the task.

\textbf{Output as Pairwise Evaluation Reference} 
Regarding $\phi_{eval}$ design, by employing the evaluation model to perform pairwise selection, we are effectively leveraging the evaluation model's inherent preference understanding of tasks. This internal signal can be obtained through simple pairwise comparisons of outputs, avoiding the need for large sample sizes to ensure scoring stability, which is typically required in score-based feedback methods.

 While we mitigate potential biases through four rounds of randomized evaluation, these biases cannot be completely eliminated~\cite{zhou2024zepo}. However, these biases do not affect the overall optimization trend because eval's feedback merely serves as a reference for the next round of optimization. The overall optimization process naturally aligns with the optimization model's task understanding, with the eval mechanism serving to validate the effectiveness of each iteration.

\section{Experiment}
\label{sec:exp}

\subsection{Experimental Setup}

\begin{table*}[t!]
\caption{Comparison of performance between conventional prompt methods and prompt generated by prompt optimization methods in five benchmarks. All methods are executed with GPT-4o-mini on the divided test set, with results averaged over three runs. \ours and \ours$^*$ use Claude-3.5-Sonnet and GPT-4o as their optimization models, respectively. The Avg. Cost refers to the averaged optimization cost.}

\label{tab:mainres}

\small
\setlength{\abovecaptionskip}{0.1cm}
\setlength{\belowcaptionskip}{-0.2cm}
\centering
\resizebox{\textwidth}{!}{
\begin{tabular}{l|ccccccc}
\hline

\hline

\hline

\hline
\multirow{2}{*}{\textbf{Method}} & \multicolumn{6}{c}{\textbf{Method Analysis}}  \\
& GPQA & AGIEval-MATH& LIAR& WSC & BBH-Navigate & Avg. Perf. & Avg. Cost(\$)\\
\hline

\hline
IO  & 38.9& 42.1& 63.5& 72.4& 91.3& 61.6& - \\
CoT~\cite{wei2022COT} & 41.6& 44.5& 65.4& 77.8& 89.7& 63.8& - \\
Rephrase~\cite{deng2023rephrase} & 40.2& 42.1& 50.5& 79.1& 93.5& 61.1& - \\
Step-back~\cite{hua2024step} & 42.4& \textbf{47.5}& 62.8& 78.7& 93.5& 65.0& - \\ 
\hline
APE~\cite{yong2023ape} & 41.1 & 44.4 & 65.9 & 80.2 & 92.5 & 64.8 & 9.07 \\
OPRO~\cite{yang2023opro} & \underline{43.3} & \underline{46.1} & \textbf{67.6} & 80.2 & 95.8 & \underline{66.6} & 4.51  \\
PromptAgent~\cite{xin2024pa} & 41.3& 41.4& 64.1& \textbf{82.7}& 95.7& 65.0 & 2.71  \\
PromptBreeder~\cite{chris2024pb} & 40.9& 45.9& 63.2& 76.7& 96.3 & 64.5 & 4.82 \\
TextGrad ~\cite{mert2024textgrad} & 40.2 & 44.4 & 65.7 & 78.0 & 91.3 & 63.9 & 13.14 \\
\rowcolor[gray]{.8}
\textbf{SPO} (ours) & \textbf{43.6}& \underline{46.1}& \underline{67.1}& \underline{82.0}& \textbf{97.2}& \textbf{66.9} & \underline{0.15}\\
\rowcolor[gray]{.8}
\textbf{SPO}$^*$ (ours) & 41.8& 45.3 & 66.9 & 81.1 & \underline{96.3} & 66.3 & \textbf{0.12} \\
\hline

\hline

\hline

\hline
\end{tabular}
}
\end{table*}

\paragraph{Datasets}
We evaluated \ours on a diverse set of tasks, including both \textit{\textbf{closed tasks}} and \textbf{\textit{open-ended tasks}}, to comprehensively assess its effectiveness.

For \textit{closed tasks}, we utilized five established benchmarks: GPQA~\cite{david2023gpqa}, AGIEval-MATH~\cite{wan2024agieval}, LIAR~\cite{william2017liar}, WSC~\cite{hector2012wsc}, and BBH-navigate~\cite{mirac2023bbh}. For WSC, LIAR, and BBH-Navigate, we sampled portions from their original datasets as test sets following~\citet{cilin2024erm}. 
For GPQA, we used the more challenging GPQA-Diamond subset as the test set, while for AGIEval-Math, we used Level 5 problems as the test set. For \textit{open-ended tasks}, we selected \textit{writing}, \textit{roleplay}, and \textit{humanities} tasks from MT-Bench~\cite{lianmin2023mtbench}. Given the limited size of the dataset, we manually constructed three validation sets for these tasks. Detailed descriptions of the datasets and the construction procedures for validation and test sets are provided in Appendix~\ref{appendix:exp}.

\paragraph{Baselines}

We evaluate \ours against two categories of methods on \textit{closed tasks}:
(1) conventional prompting approaches, comprising io (direct llm invocation), chain-of-thought, rephrase, and step-back abstract; and 
(2) automated prompt optimization methods, including APE, OPRO, PromptAgent, PromptBreeder and TextGrad with their evaluation framework setting, features detailed in Appendix \ref{appendix:po_configuration}.

For \textit{open-ended tasks} in MT-Bench~\cite{lianmin2023mtbench}, we use GPT-4o to compare outputs generated by \ours against those directly generated by the model.

\paragraph{Implementation Details}
\ours employs different models for optimization, evaluation, and execution. In the main experiments, we use Claude-3.5-Sonnet-20240620 and GPT-4o-0806 (temperature = 0.7) as optimization models, while GPT-4o-mini-0718 (temperature = 0.3 for evaluation, 0 for execution) is used for both evaluation and execution. The optimization process runs for 10 iterations with three samples per iteration, while detailed baseline implementation settings are provided in Appendix~\ref{appendix:exp}.

\paragraph{Metrics}
We evaluate performance using accuracy metrics for GPQA, LIAR, BBH-Navigate, AGIEval-MATH and WSC, following ~\citet{cilin2024erm, jon2024archon, david2023gpqa}. For MT-Bench, we report win rates of \ours compared to other methods. To assess cost-efficiency, we also measure optimization costs. To verify the stability of results, we include standard deviation metrics in Appendix ~\ref{appendix:add_experiments}.

\subsection{Experimental Results and Analysis}

\paragraph{Main Result of Closed Tasks}

As shown in Table \ref{tab:mainres}, prompts optimized by \ours outperform all conventional prompting methods on average, exceeding the best baseline by \textbf{1.9}. Meanwhile, it achieves comparable performance to ground truth-dependent prompt optimization methods across most datasets, and reaches optimal results on GPQA and BBH-navigate datasets. Specifically, \ours's superior average performance over other optimization methods demonstrates that its pairwise evaluation approach can generate more effective optimization signals compared to the tested methods relying on external reference. Furthermore, to verify the effectiveness of our method across different optimization models, we conducted experiments using GPT-4o as the optimization model, achieving an average performance of \textbf{66.3}. While slightly lower than results using Claude-3-5-Sonnet as the optimization model, this still ranks third among all compared methods.

\paragraph{Cost Analysis}

We present a comprehensive comparison of optimization costs and performance between SPO (using Claude-3.5-Sonnet and GPT-4o as optimization models) and other optimization methods in Table~\ref{tab:mainres}. While maintaining comparable performance with other ground truth-dependent prompt optimization methods, \ours requires only \textbf{1.1\%} to \textbf{5.6\%} of their optimization costs, with an average optimization cost of \textbf{0.15 \$} per dataset. The low computational cost and no need for ground truth make SPO ideal for real-world applications.

\paragraph{Ablation Study}

\begin{table}[htbp]
\caption{Performance comparison on BBH-navigate: prompting methods and \ours with different \textbf{evaluation models} (rows) and execution models (columns). The optimization model is set to Claude-3.5-Sonnet.}
\label{tab:ablation-eval}
\renewcommand\tabcolsep{3.2pt}
\renewcommand\arraystretch{1.2}
\small
\setlength{\abovecaptionskip}{0.1cm}
\setlength{\belowcaptionskip}{-0.2cm}
\centering
\resizebox{0.5\textwidth}{!}{
\begin{tabular}{l|ccc}
\hline

\hline

\hline

\hline
{\textbf{}} & \textbf{GPT-4o-mini} & \textbf{Llama3-70B} & \textbf{Claude-3-Haiku}  \\
\hline

\hline
IO & 91.3 & 82.7 & 62.2 \\
COT & 89.7 & 86.2 & 68 \\
\hline
Claude-3.5-Sonnet & 95 & 86.8 & 68.2 \\
Llama3-70B & 94.5 & 94.2 & 82.0 \\
GPT-4o-mini & \textbf{97.8}& 90.7 & 82.0 \\

\hline

\hline

\hline

\hline
\end{tabular}
}
\end{table}

\begin{table}[htbp]
\caption{Performance comparison across different \textbf{optimization models} (rows) and execution models (columns) on BBH-navigate. The evaluation model is set to GPT-4o-mini.}
\label{tab:ablation}
\renewcommand\tabcolsep{3.2pt}
\renewcommand\arraystretch{1.2}
\small
\setlength{\abovecaptionskip}{0.1cm}
\setlength{\belowcaptionskip}{-0.2cm}
\centering
\resizebox{0.5\textwidth}{!}{
\begin{tabular}{l|ccc}
\hline

\hline

\hline

\hline
{\textbf{}} & \textbf{GPT-4o-mini} & \textbf{Llama3-70B} & \textbf{Claude-3-Haiku}  \\
\hline

\hline
Claude-3.5-Sonnet & 97.2 & 86.7 & 89.7 \\
GPT-4o & 96.3 & 85.5 & 73.0 \\
GPT-4o-mini & \textbf{97.8}& 90.7 & 82.0 \\
DeepSeek-V3 & 94.7 & 83.7 & 77.2  \\
Gemma-3-12B & 97.5 & 92.0 & 69.0 \\

\hline

\hline

\hline

\hline
\end{tabular}
}
\end{table}

To evaluate the transferability of \ours across different optimization, evaluation, and execution models, we conducted ablation experiments on the BBH-Navigate dataset. The experimental results in Table \ref{tab:ablation-eval}, \ref{tab:ablation} demonstrate that \ours exhibits robust performance across different models. Notably, the best performance (\textbf{97.8}) was achieved when GPT-4o-mini was used as the optimization, execution, and evaluation model.  In terms of execution, \ours effectively improves the performance of weaker models, elevating Claude-3-Haiku from \textbf{62.2} to \textbf{89.7}, demonstrating \ours's applicability to weaker models and further expanding its potential for real-world applications.

We conduct an ablation study to investigate the impact of sample number and iteration rounds on SPO's performance using the BBH-Navigate dataset. For sample number, the performance curves of all three optimization models exhibit similar patterns: performance initially improves with increased sample number but eventually converges or declines. For iteration rounds, when we extend the optimization process to 20 rounds, we find that the actual performance of the optimal prompt selected by SPO similarly shows a pattern of initial improvement followed by degradation, which we attribute to overfitting to the sample questions. Based on these experiments, we determine the final sample number and iteration rounds for SPO, with more details shown in Appendix ~\ref{appendix:add_experiments}.

\begin{figure}[htbp]
  \centering
  \includegraphics[width=\columnwidth]{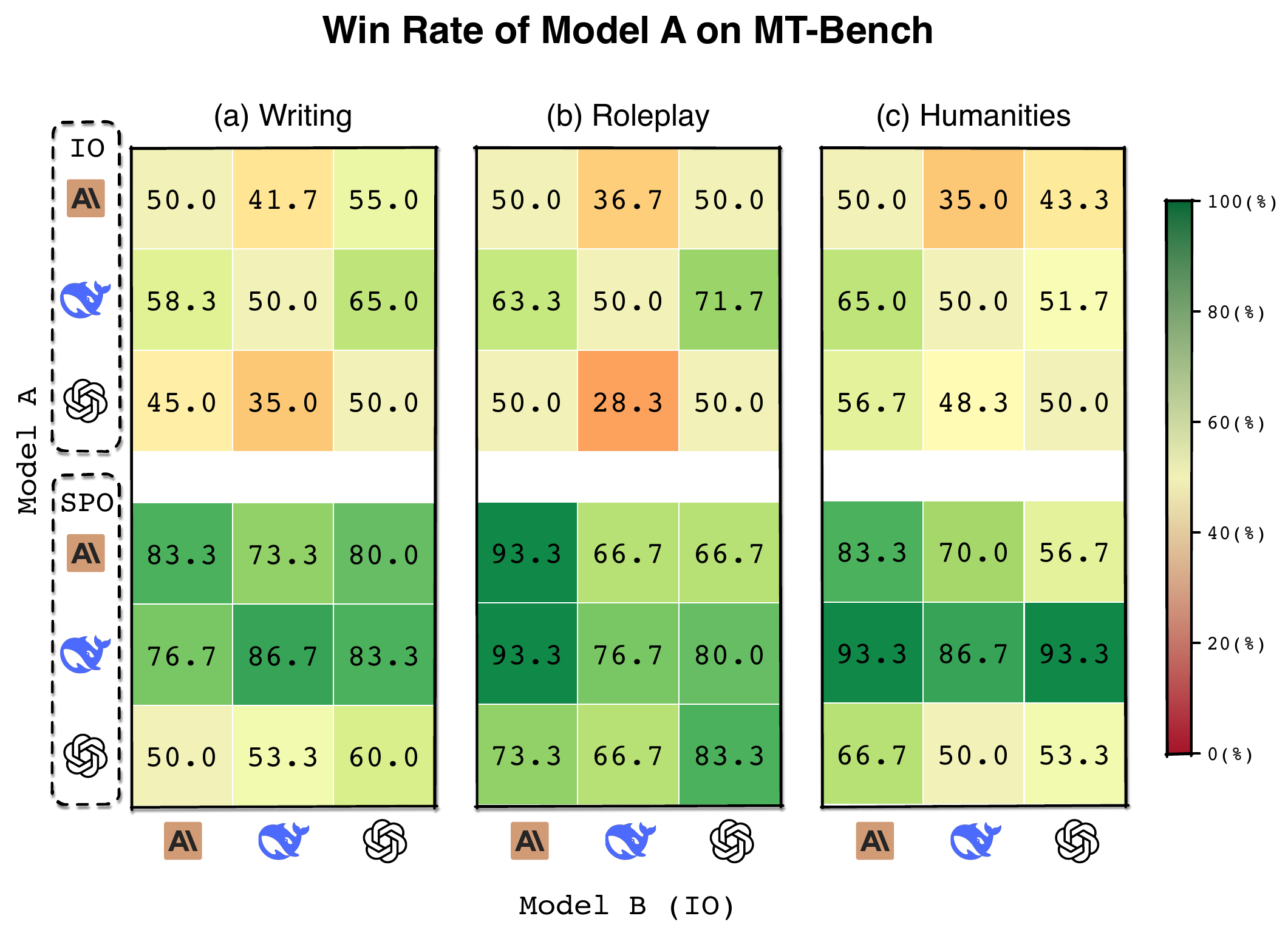}
  \vspace{-2em}
  \caption{Win rates comparison between different LLMs and \ours across three tasks. The heatmap shows pairwise win rates (\%) where each cell represents the row model's win rate against the column model. Models tested include Claude-3.5-Sonnet, DeepSeek-V3, and GPT-4o-mini. Models are evaluated both in IO (top three rows) and after SPO optimization (bottom three rows). Win rates range from 0\% to 100\%, with higher percentages indicating better performance.}
  \label{fig:mt-bench}
\end{figure}

\paragraph{Main Result of Open-ended Tasks}

To validate \ours's capability in open-ended tasks, we selected three categories from MT-Bench: ``Writing'', ``Roleplay'', and ``Humanities'' for evaluation. We use Claude-3.5-Sonnet as the optimization model, GPT-4o-mini as the evaluation model, and selected Claude-3.5-Sonnet, DeepSeek-V3, and GPT-4o-mini as execution models, conducting five iterations. Subsequently, following the evaluation methodology in~\cite{lianmin2023mtbench}, we employed GPT-4o to perform pairwise comparisons between model A and model B's output in Figure \ref{fig:mt-bench}. The experimental results shown in \ref{fig:mt-bench} demonstrate that \ours significantly improves model performance across all model configurations. 
Notably, smaller models (such as GPT-4o-mini) using optimized prompts frequently outperformed larger models in most scenarios.

\subsection{Case Study}

We present optimization results on additional open-ended tasks without datasets and \ours's optimization trajectories in the Appendix \ref{appendix:case-study}. We also provide optimal prompt across five closed tasks discoverd by \ours in the supplementary material.
Given that real-world applications often face challenges with limited datasets, we evaluate \ours's performance on tasks that lack conventional benchmarks. The experimental results, coupled with \ours's cost efficiency, demonstrate its practical value in real-world scenarios. Specifically, we demonstrate the optimization results after 10 iterations using Claude-3.5-Sonnet as the optimization model, GPT-4o-mini as the evaluation model, and Llama-3-8B as the execution model across four tasks: Advertising Design, Social Media Content, Modern Poetry Writing, and Concept Interpretation in Appendix \ref{appendix:open-ended}. Moreover, we provide a comprehensive analysis of \ours's optimization trajectory on the BBH-navigate dataset in Appendix \ref{appendix:trajectory}, presenting both successful and unsuccessful examples to offer deeper insights into the optimization process.
\section{Related Work}

\subsection{Prompt Engineering}
Research on prompting methods for large language models has primarily evolved along two directions. The first focuses on task-agnostic prompting techniques that enhance LLMs' general capabilities, including chain-of-thought \cite{wei2022COT, take2022zerocot}, single-shot reasoning enhancement \cite{deng2023rephrase, hua2024step, wang2024negativeprompt}, and output format specification \cite{zhang2024aflow, he2024doseformat, zhi2024freeformat}. The second direction addresses domain-specific prompting for tasks in code generation~\cite{sirui2024meta, tal2024alpha,10720675}, data analysis~\cite{hong2024data, liu2024surveynl2sqllargelanguage, DBLP:journals/pvldb/LiLCLT24}, question answering~\cite{DBLP:conf/emnlp/WuYSW0L24,DBLP:journals/corr/abs-2406-07815,yang2024askchartuniversalchartunderstanding}, decision-makings~\cite{zhang2024mobileexperts, guan2024voyager}, and other domains~\cite{guo2024largelanguagemodelbase,DBLP:journals/corr/abs-2404-18144,shen2024askhumansaiexploring}. As LLM applications expand to increasingly complex scenarios, manually crafting effective domain-specific prompts becomes impractical~\cite{zhang2024aflow}, motivating research in prompt optimization to systematically develop effective prompts rather than discovering general principles.

\subsection{Prompt Optimization}

The design of evaluation frameworks is crucial in Prompt Optimization (PO), determining both optimization effectiveness and computational efficiency. The evolution of evaluation mechanisms in PO has progressed from simple feedback collection to sophisticated optimization signal generation~\cite{kai2024posurvey}. Existing methods largely rely on ground truth as the evaluation source with benchmark-based numerical assessments \cite{yong2023ape, qing2024evoprompt, yang2023opro, chris2024pb, xin2024pa, omar2024dspy}, often requiring substantial iterations and samples, while others use LLM-as-a-judge~\cite{lianmin2023mtbench} to generate detailed textual feedback \cite{cilin2024erm, mert2024textgrad, yu2024stargo, wang2024semantic, reid2023protegi, ya2025ttpo}, still depending on ground truth data. Some approaches focus on human preferences through evaluation rules or direct feedback \cite{yong2024promst, lin2024apohf}, while others explore alternative criteria like output consistency \cite{xuan2024glape}.

In contrast to these approaches, \ours introduces a novel evaluation paradigm that eliminates dependency on external references while maintaining efficiency. By leveraging only model outputs through pairwise comparisons, \ours achieves robust evaluation without requiring ground truth, human feedback, or extensive sampling, making it particularly suitable for real-world applications.

\section{Conclusion}

This paper addresses a key challenge in prompt optimization: dependency on external references. We introduce Self-Supervised Prompt Optimization (\ours), a framework that overcomes this dependency while achieving remarkable cost-efficiency at only \textbf{\$0.15} per dataset. Drawing inspiration from self-supervised learning, \ours innovatively constructs evaluation and optimization signals through pairwise comparisons of model outputs, enabling reference-free optimization without compromising effectiveness. Our evaluation shows \ours achieves state-of-the-art performance across diverse tasks while requiring only 1.1\%-5.6\% of existing methods' costs. Success on both benchmarks and real-world applications validates \ours's effectiveness and generalization capabilities, making prompt optimization more accessible and practical across various domains.

\clearpage
\section*{Limitations}

Despite \ours's promising performance and cost efficiency, several limitations should be acknowledged. First, while our method eliminates dependence on external references, its effectiveness remains contingent on the evaluation model's ability to accurately assess output quality, which may introduce biases when the evaluation model lacks sufficient understanding of specialized domains. Second, our approach currently focuses on optimizing prompts for a single model; future work should explore cross-model prompt optimization to enhance transferability across different LLMs with varying capabilities.

\section*{Ethical Consideration} 
\ours offers significant advancements in prompt engineering for LLMs, offering benefits such as democratized access, reduced costs, and improved performance across various tasks. However, it also carries risks, including potential bias amplification, misuse of harmful content generation, and over-reliance on LLMs. 

\bibliography{cited}

\clearpage
\appendix

\section{Appendix}

\setcounter{table}{0}  
\setcounter{figure}{0}
\setcounter{algorithm}{0}
\renewcommand{\thetable}{A\arabic{table}}
\renewcommand{\thefigure}{A\arabic{figure}}

\subsection{Detailed Prompts of \ours}
\label{appendix:prompt}

In this section, we present the Meta Prompt used for iteration. It should be noted that here we have only used the simplest and most straightforward Prompt. There is still room for improvement by optimizing the following Meta Prompt for specific domains.

\begin{tcolorbox}[title={\textbf{\small Optimize Function's Prompt}}, boxrule=2pt, arc=0mm, breakable]
\begin{minted}[fontsize=\scriptsize, breaklines, breakanywhere, frame=lines, framesep=2mm, tabsize=4, style=vs, autogobble]{python}
PROMPT = """You are building a prompt to address user requirement.Based on the given prompt, please reconstruct and optimize it. You can add, modify, or delete prompts. Please include a single modification in XML tags in your reply. During the optimization, you can incorporate any thinking models.
This is a prompt that performed excellently in a previous iteration. You must make further optimizations and improvements based on this prompt. The modified prompt must differ from the provided example.

requirements:
```
{requirements}
```

reference prompt:
```
{prompt}
```

The execution result of this reference prompt is(some cases):
```
{answers}
```

Provide your analysis, optimization points, and the complete optimized prompt using the following XML format:

<analyse>Analyze what drawbacks exist in the results produced by the reference prompt and how to improve them.</analyse>
<modification>Summarize the key points for improvement in one sentence</modification>
<prompt>Provide the complete optimized prompt</prompt>"""
\end{minted}
\end{tcolorbox}

This prompt template guides LLMs to iteratively improve existing prompts through structured XML analysis. It requires identifying weaknesses in reference prompt outputs, proposing modifications, and generating optimized versions. The template emphasizes incremental improvements while maintaining requirement compliance.

\begin{tcolorbox}[title={\textbf{\small Evaluate Function's Prompt}}, boxrule=2pt, arc=0mm, breakable]\begin{minted}[fontsize=\scriptsize, breaklines, breakanywhere, frame=lines, framesep=2mm, tabsize=4, style=vs, autogobble]{python}
PROMPT = """Based on the original requirements, evaluate the two responses, A and B, and determine which one better meets the requirements. If a reference requirement is provided, strictly follow the format/content of the reference requirement.

# Requirement
{requirement}

# A
{Answer_A}

# B
{Answer_B}


Provide your analysis and the choice you believe is better, using XML tags to encapsulate your response.

<analyse>Some analysis</analyse>
<choose>A/B (the better answer in your opinion)</choose>"""
\end{minted}
\end{tcolorbox}

The evaluation template uses comparative analysis to assess response quality. It requires XML-formatted reasoning that analyzes strengths/weaknesses of two responses (A/B) against requirements, followed by a definitive choice. 

\subsection{Detailed Prompt Template of Iteration Start}

\begin{tcolorbox}[title={\textbf{\small Iteration template on the BBH-navigate dataset}}, boxrule=2pt, arc=0mm, breakable]\begin{minted}[fontsize=\scriptsize, breaklines, breakanywhere, frame=lines, framesep=2mm, tabsize=4, style=vs, autogobble]{yaml}

prompt: |
  Please think step by step.
  Ensure the response concludes with the answer in the XML format: 
  <answer>[Yes or No]</answer>.

requirements: |
  Must put the final answer at the end with XML. (<answer>(Yes or No)</answer>,such as <answer>Yes</answer>)
  The provided prompt needs to adapt to all current types of questions.

faq:
  - question: |
        If you follow these instructions, do you return to the starting point? Always face forward. Take 7 steps left. Take 2 steps backward. Take 7 steps backward. Take 7 steps backward. Take 3 steps forward.
        Options:
        - Yes
        - No

    answer: |
        A lot of thinking and analysis processes.
        ...
        Final Answer:
        <answer>(Yes or No)</answer>

  - question: |
        If you follow these instructions, do you return to the starting point? Always face forward. Take 6 steps backward. Take 8 steps left. Take 3 steps right. Take 7 steps forward. Take 3 steps right. Take 9 steps right. Take 1 step backward. Take 7 steps left.
        Options:
        - Yes
        - No

    answer: |
        A lot of thinking and analysis processes.
        ...
        Final Answer:
        <answer>(Yes or No)</answer>

  - question: |
        If you follow these instructions, do you return to the starting point? Turn left. Turn left. Take 6 steps. Take 3 steps. Turn around. Take 1 step. Take 3 steps. Take 5 steps.
        Options:
        - Yes
        - No

    answer: |
        A lot of thinking and analysis processes.
        ...
        Final Answer:
        <answer>(Yes or No)</answer>
\end{minted}
\end{tcolorbox}

This YAML file demonstrates the initial configuration for our approach to iterating on the BBH-navigate task. By configuring a simple initial Prompt and requirements, along with three specific questions, iterative optimization can be performed. It should be noted that the content shown here is the complete content of the file, and the content in the answer section is not the actual answer but serves as a reference for the thought process and correct output format.

\subsection{Experiment Details}
\label{appendix:exp}

\subsubsection{Tasks and Data Details}


\begin{table}[htbp]
\caption{Dataset sizes and data splits.}
\label{tab:dataset}
\renewcommand\tabcolsep{3.2pt}
\renewcommand\arraystretch{1.2}
\small
\setlength{\abovecaptionskip}{0.1cm}
\setlength{\belowcaptionskip}{-0.2cm}
\centering
\begin{tabular}{l|cc}
\hline

\hline

\hline

\hline
\textbf{Dataset Name} & \textbf{Test} & \textbf{Train\&Validate} \\
\hline

\hline
LIAR & 461 & 3681  \\
BBH-Navigate & 200 & 50  \\
WSC & 150 & 50  \\
AGIEval-MATH & 256 & 232 \\
GPQA & 198 & 250  \\
\hline
MT-bench & 80 & 0  \\

\hline

\hline

\hline

\hline
\end{tabular}
\end{table}

\textbf{LIAR}
LIAR~\cite{william2017liar} is an English fake news detection dataset consisting of 4,000 statements, each accompanied by contextual information and lie labels. For our experiments, we sampled portions from the original dataset as test sets following~\citet{cilin2024erm}.
\textbf{BBH-Navigate}
BBH-Navigate~\cite{mirac2023bbh} is a task from the BIG-bench Hard dataset, a subset of the BIG Bench dataset. This task focuses on navigation reasoning, requiring the model to determine whether an agent, after following a series of navigation steps, returns to its starting point. For our experiments, we employed random sampling (seed=42) to obtain 200/25/25 test/train/validation splits.
\textbf{WSC}
The Winograd Schema Challenge (WSC)~\cite{hector2012wsc} is a benchmark designed to evaluate a system's ability to perform commonsense reasoning by resolving pronoun references in context. For our experiments, we sampled portions from the original dataset as test sets following~\citet{cilin2024erm}.
\textbf{AGIEval-MATH}
AGIEval-MATH~\cite{wan2024agieval} is a subset of the AGIEval benchmark, focusing on mathematical problem-solving tasks. It includes a variety of math problems designed to assess reasoning and computational abilities. For our experiments, we used Level 5 problems as the test set and Level 4 problems as the training and validation set.
\textbf{GPQA}
GPQA~\cite{david2023gpqa} is a dataset designed to evaluate the performance of language models on graduate-level questions across multiple disciplines, including biology, physics, and chemistry. For our experiments, we utilized the GPQA-Diamond subset as the test set, while constructing our training and validation set from questions that are exclusive to GPQA-main (i.e., those present in GPQA-main but absent from GPQA-Diamond).
\textbf{MT-bench}
MT-bench~\cite{lianmin2023mtbench} is a multi-task benchmark designed to evaluate the generalization capabilities of language models across diverse tasks, including text classification, summarization, and question answering. For our experiments, we selected \textit{writing}, \textit{roleplay}, and \textit{humanities} tasks from MT-Bench. These validation questions are provided in the supplementary materials.

\subsubsection{Configuration}
\label{appendix:po_configuration}

In our experiments, we configured different optimization frameworks to align their optimization costs as much as possible. These frameworks generally allow setting some parameters to adjust optimization costs, including the number of iterations and the number of prompts generated per iteration. Specifically, we compare the features of prompt optimization methods we used in Table ~\ref{tab:appendix:features}.

\begin{table}[htbp]
\caption{Comparison of evaluation frameworks across different prompt optimization methods. OvG denotes evaluation against ground truth, while OvO represents output-vs-output comparison. Methods are categorized by their evaluation source and method.}
\label{tab:appendix:features}
\renewcommand\tabcolsep{3.2pt}
\renewcommand\arraystretch{1.2}
\small
\setlength{\abovecaptionskip}{0.1cm}
\setlength{\belowcaptionskip}{-0.2cm}
\centering
\begin{tabular}{l|cc}
\hline

\hline

\hline

\hline
{\textbf{Method}} & \textbf{Evaluation Source} & \textbf{Evaluation Method}   \\
\hline

\hline
APE & OvG & Benchmark  \\
OPRO & OvG & Benchmark  \\
PromptAgent & OvG & Benchmark \\
PromptBreeder & OvG & Benchmark \\
TextGrad & OvG & LLM as a judge \\
\hline
\rowcolor[gray]{.8}
\ours & \textbf{OvO} & LLM as a judge \\

\hline

\hline

\hline

\hline
\end{tabular}
\end{table}

\textbf{APE}
APE employs a three-round iterative optimization process, selecting the top 10\% (ratio=0.1) performing prompts from the current pool as elite prompts in each round. To maintain diversity and size of the prompt pool, variant sampling is used to mutate these elite prompts, keeping the total number of prompts at 50. Following the setting in original paper~\cite{yong2023ape}, the optimization process does not incorporate specific sample execution results to guide LLM prompt optimization. Instead, performance scores are obtained by evaluating prompts on the entire training set.

\textbf{OPRO}
OPRO uses a 10-round iterative optimization process, generating 10 candidate prompts per round. OPRO evaluates prompt performance on the complete training set and filters based on evaluation scores. OPRO doesn't maintain a fixed-size prompt pool but directly generates new candidates based on the current best prompt in each round. The optimization direction is guided through performance evaluation on the full training data.

\textbf{PromptAgent}
Except for the Liar dataset, on which we sampled 150 data for both training and validation, all other datasets follow the sizes specified in Table~\ref{tab:dataset}. PromptAgent uses a Monte Carlo Tree Search (MCTS) framework to optimize prompts. It starts with an initial prompt and generates new candidates based on model error feedback. The process is guided by evaluations with benchmark on a sampled training set to identify high-reward paths for improved task performance. Finally, we select the top 5 prompts that perform best on the validation set for testing and choose the optimal one. Key parameters of MCTS include an expand width of 3, a depth limit of 8, and 12 iterations. 

\textbf{PromptBreeder}
In our implementation of PromptBreeder, we configure the system with 5 mutation prompts and 5 thinking styles for initialization. The evolution process runs for 20 generations, with 20 evaluations performed on randomly sampled training examples in each generation. The optimization model defaults to Claude-3.5-Sonnet and the execution model defaults to GPT-4o-mini.

\textbf{TextGrad}
Except for the Liar dataset where the Train\&Validate set is reduced to 50 samples, all other datasets follow the sizes specified in Table~\ref{tab:dataset}. TextGrad employs a three-epoch optimization process with three steps per epoch (epoch=3, steps=3), using a batch size of 3 for stochastic gradient descent. In each step, TextGrad generates gradients through back-propagation of feedback from the optimizer LLM (Claude-3.5-Sonnet) to update the system prompt. The framework maintains a validation-based reversion mechanism - if the updated prompt performs worse on the validation set compared to the previous iteration, the update is rejected and the prompt reverts to its previous state. The optimization process is guided by evaluating prompts using Claude-3.5-Sonnet as the evaluation LLM, while the actual task execution uses GPT-4o-mini as the execution LLM. Our experimental configuration follows the prompt optimization setting provided in the official TextGrad repository~\cite{mert2024textgrad}.

\textbf{\ours}
\ours conducts optimization through 10 iterations per task, randomly selecting 3 questions (without answers) from the pre-split Train\&Validate dataset for each iteration. The optimization model defaults to Claude-3.5-Sonnet, the evaluation model defaults to GPT-4o-mini, and the execution model defaults to GPT-4o-mini. Notably, \ours demonstrates effective prompt optimization using only questions without ground truth answers, validating its capability.

\subsubsection{Baseline Prompt}

In this section, we provide the Baseline Prompts for comparison. Note that for all Prompt Optimization work requiring initial iteration prompts, we consistently provide the COT Prompt shown below.

\begin{tcolorbox}[title={\textbf{\small IO Prompt}}, boxrule=2pt, arc=0mm, breakable]
\begin{minted}[fontsize=\scriptsize, breaklines, breakanywhere, frame=lines, framesep=2mm, tabsize=4, style=vs, autogobble]{markdown}

Ensure the response concludes with the answer in the format: 
<answer>answer</answer>

\end{minted}
\end{tcolorbox}

\begin{tcolorbox}[title={\textbf{\small COT Prompt}}, boxrule=2pt, arc=0mm, breakable]
\begin{minted}[fontsize=\scriptsize, breaklines, breakanywhere, frame=lines, framesep=2mm, tabsize=4, style=vs, autogobble]{markdown}

Please think step by step.
Ensure the response concludes with the answer in the format:
<answer>answer</answer>.

\end{minted}
\end{tcolorbox}

\begin{tcolorbox}[title={\textbf{\small Step-back Prompt}}, boxrule=2pt, arc=0mm, breakable]
\begin{minted}[fontsize=\scriptsize, breaklines, breakanywhere, frame=lines, framesep=2mm, tabsize=4, style=vs, autogobble]{markdown}

Please first think about the principles involved in solving this task which could be helpful.
And Then provide a solution step by step for this question.
Ensure the response concludes with the answer in the format:
<answer>answer</answer>.

\end{minted}
\end{tcolorbox}

\begin{tcolorbox}[title={\textbf{\small Rephrase Prompt}}, boxrule=2pt, arc=0mm, breakable]
\begin{minted}[fontsize=\scriptsize, breaklines, breakanywhere, frame=lines, framesep=2mm, tabsize=4, style=vs, autogobble]{markdown}

Please rephrase the question in a way that is easier to understand, minimizing ambiguity and considering edge cases.
And Then provide a solution step by step for the question.
Ensure the response concludes with the answer in the format:
<answer>answer</answer>.

\end{minted}
\end{tcolorbox}

\subsubsection{Prompt Optimized by \ours}

In this section, we present the optimized prompts obtained from our main experiments, where Claude-3.5-Sonnet serves as the optimization model, and GPT-4o-mini serves as both the evaluation and execution model.

\begin{tcolorbox}[title={\textbf{\small GPQA Prompt}}, boxrule=2pt, arc=0mm, breakable]
\begin{minted}[fontsize=\scriptsize, breaklines, breakanywhere, frame=lines, framesep=2mm, tabsize=4, style=vs, autogobble]{markdown}

Follow these guidelines to answer questions efficiently and effectively:

1. Carefully read the entire question, identifying all relevant information and key concepts.
2. Choose the most appropriate problem-solving approach based on the question type.
3. Solve the problem using these steps:
   a. State any relevant formulas, principles, or assumptions
   b. Show all necessary calculations or conceptual analysis
   c. Evaluate all answer options, explaining why incorrect options are wrong when relevant
4. Structure your response as follows:

[Analysis]
- Briefly state the main problem and key information (2-3 sentences max)

[Solution]
- Show your work step-by-step, including all relevant calculations and reasoning
- For conceptual questions, provide a clear, logical explanation

[Conclusion]
- State the final answer in one clear sentence
- Briefly explain why this answer is correct and others are incorrect (if applicable)

<answer>[One letter representing the correct option]</answer>

Adapt this structure as needed for different question types, prioritizing clarity and conciseness. Ensure that your response addresses all aspects of the question and demonstrates a clear problem-solving process.

\end{minted}
\end{tcolorbox}

\begin{tcolorbox}[title={\textbf{\small BBH-Navigate Prompt}}, boxrule=2pt, arc=0mm, breakable]
\begin{minted}[fontsize=\scriptsize, breaklines, breakanywhere, frame=lines, framesep=2mm, tabsize=4, style=vs, autogobble]{markdown}

Follow these steps to analyze the given instructions:

1. State the initial conditions:
   - Starting point: (0, 0)
   - Initial direction: positive x-axis (unless specified otherwise)

2. Use a coordinate system:
   - x-axis: left (-) and right (+)
   - y-axis: backward (-) and forward (+)

3. Analyze each step:
   - For ambiguous instructions (e.g., "Take X steps" without direction), assume forward movement
   - Update coordinates after each step
   - Briefly explain any assumptions made

4. After analyzing all steps:
   - Summarize total movement in each direction
   - State the final position

5. Compare final position to starting point:
   - Calculate the distance from (0, 0)

6. Provide concise reasoning, labeled as "Reasoning:"
   - Explain key movements and their impact on position
   - Justify your conclusion

7. State your final answer, labeled as "Final Answer:"

Conclude your response with the answer in this XML format:
<answer>[Yes or No]</answer>

Ensure your analysis adapts to all question types, handling both specific and ambiguous instructions.

\end{minted}
\end{tcolorbox}

\begin{tcolorbox}[title={\textbf{\small LIAR Prompt}}, boxrule=2pt, arc=0mm, breakable]
\begin{minted}[fontsize=\scriptsize, breaklines, breakanywhere, frame=lines, framesep=2mm, tabsize=4, style=vs, autogobble]{markdown}

Analyze the given statement(s) carefully, following these steps for each question:

1. Consider the statement, speaker's background (if provided), and context.
2. Research and cite relevant facts and data related to the claim.
3. Evaluate the claim's validity based on available evidence.
4. Consider potential biases or motivations of the speaker.

For each claim, structure your analysis as follows:

CLAIM: [Restate the claim]
CONTEXT: [Summarize speaker background and context]
FACT-CHECK:
- [Key fact 1 with source]
- [Key fact 2 with source]
- [Add more as needed]
ANALYSIS: [Provide a brief analysis of the claim's validity]
RATING: [Use the following scale]
1 = False
2 = Mostly False
3 = Mixed / Unclear
4 = Mostly True
5 = True
UNCERTAINTY: [Low / Medium / High - based on the quality and quantity of available evidence]
SUMMARY: [One-sentence conclusion about the claim's validity]

If there is insufficient information to make a determination, clearly state this and explain what additional information would be needed.

Conclude each analysis with a final true/false judgment in XML tags:
<answer>[Yes or No]</answer>

After analyzing all claims, provide an overall summary of your findings.

Ensure your response is objective, based on verifiable information, and cites sources for all key facts used in the analysis.

\end{minted}
\end{tcolorbox}

\begin{tcolorbox}[title={\textbf{\small AGIEval-MATH Prompt}}, boxrule=2pt, arc=0mm, breakable]
\begin{minted}[fontsize=\scriptsize, breaklines, breakanywhere, frame=lines, framesep=2mm, tabsize=4, style=vs, autogobble]{markdown}

1. Quickly identify key information
2. Devise fastest solution strategy
3. Execute minimal calculations
4. Double-check answer

Rules:
- Use shortcuts and estimation when possible
- Skip detailed explanations
- Omit units in final answer
- Always present final answer as: \boxed{}

Prioritize speed and accuracy. Manage time efficiently. Aim for shortest valid solution. Verify answer format before submission. If stuck, make educated guess rather than spending too much time.

\end{minted}
\end{tcolorbox}

\begin{tcolorbox}[title={\textbf{\small WSC Prompt}}, boxrule=2pt, arc=0mm, breakable]
\begin{minted}[fontsize=\scriptsize, breaklines, breakanywhere, frame=lines, framesep=2mm, tabsize=4, style=vs, autogobble]{markdown}

Analyze the question using these steps:
1. Identify key elements and relationships.
2. Consider multiple interpretations.
3. Evaluate evidence for each possibility.
4. Draw a logical conclusion.
5. Summarize your reasoning.
Conclude with:
Final Answer:
<answer>[Single letter option]</answer>

\end{minted}
\end{tcolorbox}

\subsection{Addtional Experiments}
\label{appendix:add_experiments}

\subsubsection{Stability of \ours}

Check the results in Table ~\ref{tab:appendix:sta}.

\begin{table}[htbp]
\caption{Performance stability analysis of \ours. 
Standard deviation values are reported for the main experiments, where \ours and \ours$^*$ use Claude-3.5-Sonnet and GPT-4o as their optimization models, respectively. }

\label{tab:appendix:sta}
\renewcommand\tabcolsep{3.2pt}
\renewcommand\arraystretch{1.2}
\small
\setlength{\abovecaptionskip}{0.1cm}
\setlength{\belowcaptionskip}{-0.2cm}
\centering
\begin{tabular}{l|cc}
\hline

\hline

\hline

\hline
Datasets & \textbf{SPO} & \textbf{SPO}$^*$  \\
\hline

\hline
GPQA& 43.6 ± 0.31 & 41.8 ± 2.28  \\
AGIEval-MATH & 46.1 ± 0.69 & 45.3 ± 1.43  \\
LIAR & 67.1 ± 0.59 & 66.9 ± 1.20 \\
WSC & 82.0 ± 1.15 & 81.1 ± 1.01 \\
BBH-Navigate & 97.2 ± 0.29 & 96.3 ± 1.04 \\

\hline

\hline

\hline

\hline
\end{tabular}
\end{table}

\subsubsection{Ablation Study on Sample Number}

\begin{figure}[htbp]
  \centering
  \includegraphics[width=\columnwidth]{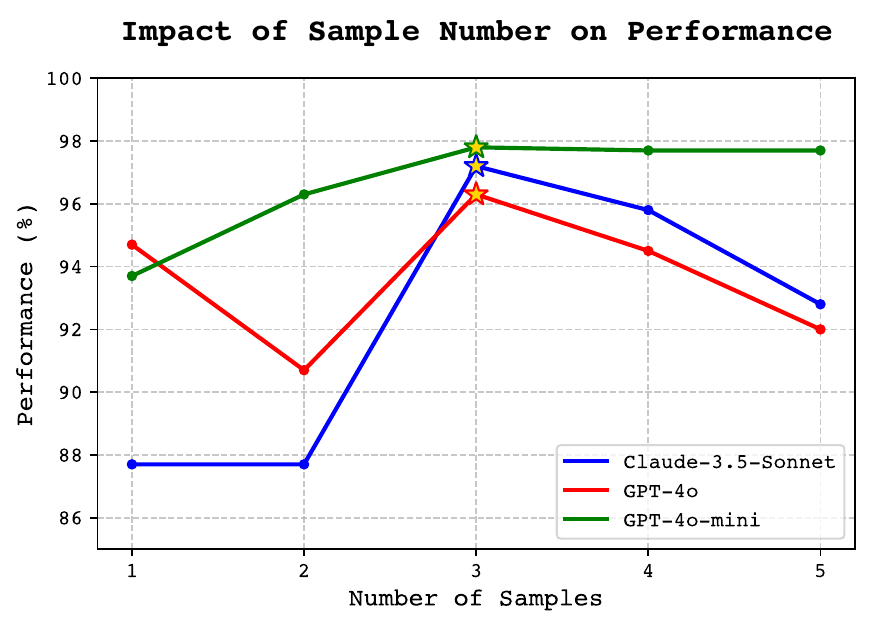}
  \vspace{-2em}
  \caption{Impact of sample number on performance across different optimization models on BBH-Navigate dataset. We evaluate three optimization models: GPT-4o-mini, GPT-4o, and Claude-3.5-Sonnet. The results demonstrate an inverted U-shaped relationship between sample number and performance.}
  \label{appendix:fig:sample}
\end{figure}

The performance curves (shown in Figure ~\ref{appendix:fig:sample})of all three optimization models exhibit similar patterns: performance initially improves with increased sample number but eventually  converges or decline. This phenomenon can be attributed to two factors: insufficient samples lead to overfitting in prompt optimization, while excessive samples not only increase computational costs but also result in longer context for the evaluation model, potentially degrading assessment quality. Based on extensive experiments, we determine that a sample size of 3 achieves the optimal balance between cost-efficiency and performance.

\subsubsection{Ablation Study on Iteration Rounds}

\begin{figure}[htbp]
  \centering
  \includegraphics[width=\columnwidth]{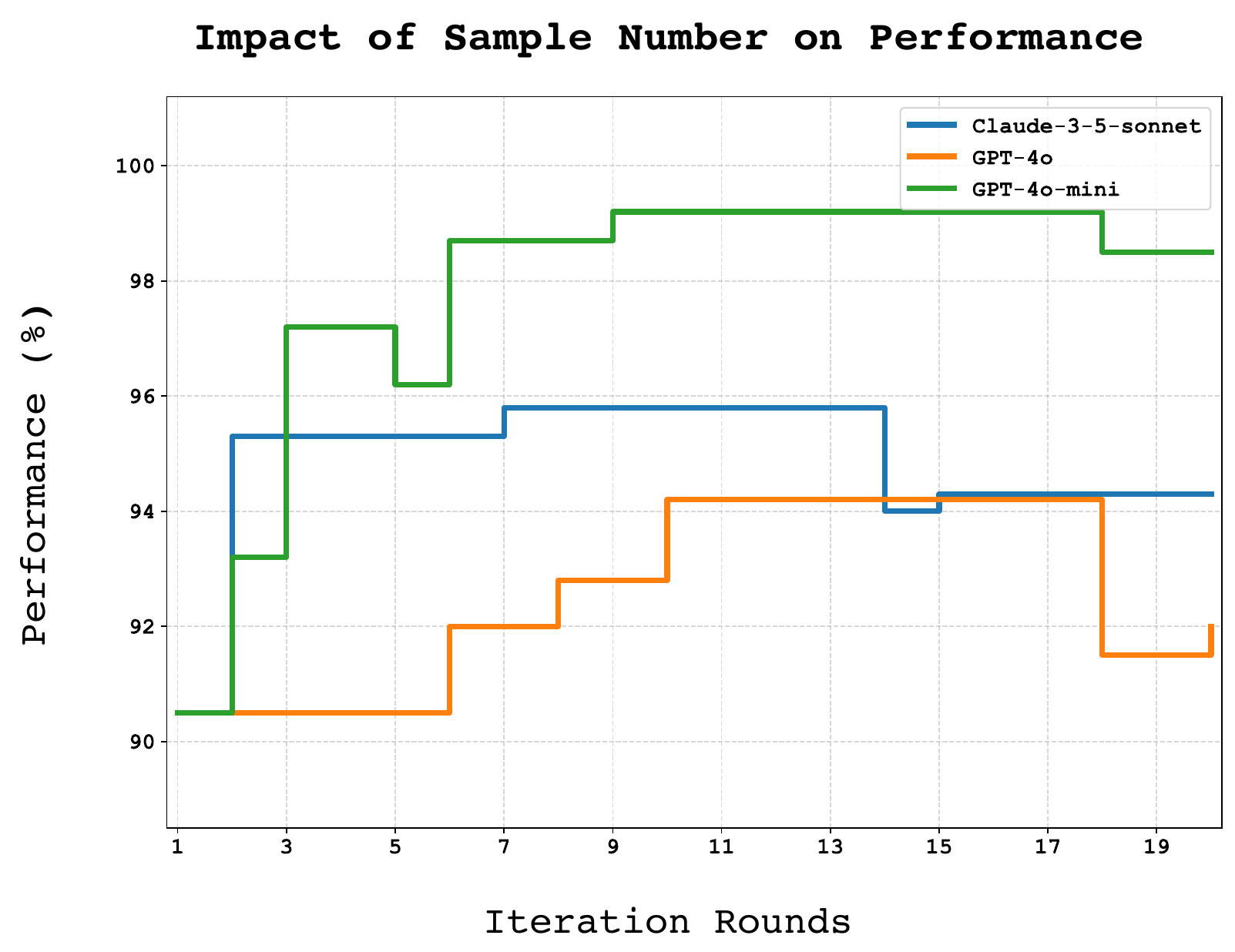}
  \vspace{-2em}
  \caption{Impact of iteration rounds on performance across different optimization models on BBH-Navigate dataset. We evaluate three optimization models: GPT-4o-mini, GPT-4o, and Claude-3.5-Sonnet.}
  \label{appendix:fig:rounds}
\end{figure}

The choice of 10 iterations is based on comprehensive empirical analysis. Taking BBH task as an example, we conducted extended optimization experiments (up to 20 iterations) using three different optimization models (Claude-3.5-Sonnet, GPT-4o, GPT-4o-mini), with GPT-4o-mini as both evaluation and execution model. The performance of the best prompts (as judged by the evaluation model) at different iteration steps is shown below (mean ± std over 3 runs):

As demonstrated by the results (plotted in Figure ~\ref{appendix:fig:rounds}), SPO typically discovers the optimal prompts within 10 iterations. Further iterations not only increase computational costs but may lead to question-specific overfitting, potentially degrading overall performance. For instance, with Claude-3.5-Sonnet, the performance peaks at iteration 7 (95.8\%) and gradually decreases afterwards. Similar patterns are observed across different optimization models.

\subsection{Case Study}
\label{appendix:case-study}
\subsubsection{Optimization Trajectory}
\label{appendix:trajectory}
We present the prompt optimization trajectory of GPT-4o-mini on the BBH-navigate dataset, where it serves as the optimization model, evaluation model, and execution model. This includes whether each iteration was successful relative to the best prompt at that time, as well as the corresponding prompt content.

\newcommand{\successbox}[2]{%
    \begin{tcolorbox}[
        title={\textbf{\small Iteration #1 - Success}},
        colframe=green!40!black!90!white,    
        colback=green!10!white!95!black,     
        coltitle=white,
        colbacktitle=green!40!black!90!white,
        boxrule=2pt,
        arc=0mm,
        left=8pt,
        right=8pt,
        top=8pt,
        bottom=8pt,
        fonttitle=\sffamily\bfseries,
        verbatim
    ]
    #2
    \end{tcolorbox}
}

\newcommand{\failedbox}[2]{%
    \begin{tcolorbox}[
        title={\textbf{\small Iteration #1 - Failed}},
        colframe=red!40!black!90!white,      
        colback=red!10!white!95!black,       
        coltitle=white,
        colbacktitle=red!40!black!90!white,  
        boxrule=2pt,
        arc=0mm,
        left=8pt,
        right=8pt,
        top=8pt,
        bottom=8pt,
        fonttitle=\sffamily\bfseries,
        verbatim
    ]
    #2
    \end{tcolorbox}
}

\successbox{0}{
Please think step by step.
Ensure the response concludes with the answer in the following format:
<answer>[Yes or No]</answer>
}

\failedbox{1}{
Please analyze the following question step by step, ensuring clarity in your reasoning. At the conclusion of your analysis, provide the final answer in the specified XML format: <answer>(Yes or No)</answer>.<br/><br/>Question: [Insert Question Here]
}

\failedbox{2}{
Please analyze the given instructions step by step and summarize your reasoning clearly. Ensure the response concludes with the answer in the XML format: 
<answer>(Yes or No)</answer>.
}

\failedbox{3}{
Please analyze the provided steps thoroughly and document each movement. Ensure your reasoning is clear and logical throughout. Conclude with the final answer in the following XML format: <answer>(Yes or No)</answer>.
}

\successbox{4}{
Please approach the question with a clear and logical reasoning process, breaking it down step by step. Ensure to summarize your findings before stating the final answer clearly in this XML format: <answer>[Yes or No]</answer>. 

For example, if the question is about returning to the starting point after a series of movements, detail each movement, updating your position accordingly, and conclude whether you have returned to the starting point or not.

Final Answer:
<answer>(Yes or No)</answer>
}

\successbox{5}{
Please analyze the question thoroughly by breaking it down step by step. Clearly outline each movement or action, updating your position or state accordingly. Conclude with a summary of your findings to determine whether you return to the starting point or achieve the stated outcome. Finally, present the answer distinctly in this XML format: <answer>[Yes or No]</answer>.

For example, if asked about returning to the starting point after several movements, document each movement clearly, track your position, and specify if you have returned to the starting point.

Final Answer:
<answer>(Yes or No)</answer>
}

\successbox{6}{
Please analyze the question methodically, breaking it down into clear, sequential movements. Define the starting point and the direction of each movement, updating the position accordingly after each step. At the end, summarize the journey and explicitly state whether the final position returns to the starting point or achieves the desired outcome. Present the final answer distinctly in this XML format: <answer>[Yes or No]</answer>.

For example, if asked about returning to the starting point after a series of movements, document each step clearly, track your position, and specify if you have returned to the starting point.

Final Answer:
<answer>(Yes or No)</answer>
}

\failedbox{7}{
Please evaluate the question step-by-step, clearly outlining each movement and tracking the resulting position. At the conclusion of your analysis, summarize the journey and state whether the final position returns to the starting point or meets the desired criteria. Ensure the final answer is presented distinctly in this XML format: <answer>[Yes or No]</answer>.

For instance, if asked about returning to the starting point after a series of movements, succinctly document each step, track your position, and indicate if you have returned to the starting point.

Final Answer:
<answer>(Yes or No)</answer>
}

\successbox{8}{
Please methodically analyze the question by breaking it down into concise movements. Start from a defined position and track each step's impact on your location. After detailing each movement, summarize the overall journey and explicitly state whether the final position returns to the starting point or reaches the desired outcome. Present the final answer distinctly in the following XML format: <answer>Yes or No</answer>.

For example, if asked whether following a series of movements returns you to the starting point, document each step clearly, track your position accurately, and specify the result.

Final Answer:
<answer>(Yes or No)</answer>
}

\failedbox{9}{
Please begin by outlining your movements systematically. Start from a defined position and detail each step you take, including the direction and distance moved. After completing your movements, summarize the overall journey, and clearly state whether your final position returns to the starting point or achieves the desired outcome. Make sure to present your movements alongside the final answer in the following XML format: <answer>Yes or No</answer>.

For example, if asked whether following a series of movements returns you to the starting point, you should document each step, track your position accurately, and specify the result. 

Final Answer:
<answer>(Yes or No)</answer>
}

\successbox{10}{
Please thoroughly analyze the provided movement instructions by documenting each step clearly. Begin by defining your initial position and systematically track your movements, updating your position after each step. After completing the movements, summarize your journey and determine if your final position coincides with the starting point or achieves the desired outcome. Clearly indicate your final answer using the XML format: <answer>Yes or No</answer>.

For instance, if the question asks whether a sequence of movements returns you to the starting point, methodically outline each step, track your position accurately, and conclude with the result.

Final Answer:
<answer>(Yes or No)</answer>
}


The success/failure boxes demonstrate \ours' optimization process through 10 iterations of a BBH-navigation prompt. Successful versions (green) gradually improve the prompt to elicit better thought processes, while failed attempts (red) may lead to oversimplified or redundant thinking due to suboptimal prompts. This trajectory shows the system's trial-and-error refinement mechanism. 

\begin{tcolorbox}[title={\textbf{\small Optimize Function's Prompt}}, boxrule=2pt, arc=0mm, breakable]
\begin{minted}[fontsize=\scriptsize, breaklines, breakanywhere, frame=lines, framesep=2mm, tabsize=4, style=vs, autogobble]{python}

PROMPT = """Please think step by step.
Ensure the response concludes with the answer in the XML format: 
<answer>[Yes or No]</answer>."""

Question = """If you follow these instructions, do you return to the starting point? Always face forward. Take 1 step right. Take 3 steps left. Take 2 steps right.
Options:
- Yes
- No"""

Answer = """Let's analyze the steps one by one:
1. **Starting Point**: We begin at the starting point, facing forward.
2. **Step 1**: Take 1 step right. Now we are 1 step to the right of the starting point.
3. **Step 2**: Take 3 steps left. From our current position (1 step right), moving 3 steps left means:
   - 1 step left brings us back to the starting point.
   - 2 steps left takes us 1 step to the left of the starting point.
   - 3 steps left takes us 2 steps to the left of the starting point.
4. **Step 3**: Take 2 steps right. From our current position (2 steps left), moving 2 steps right means:
   - 1 step right brings us back to 1 step left of the starting point.
   - 2 steps right brings us back to the starting point.
After completing all the steps, we end up back at the starting point.
Therefore, the answer is:
<answer>Yes</answer>"""


\end{minted}
\end{tcolorbox}

\begin{tcolorbox}[title={\textbf{\small Optimize Function's Prompt}}, boxrule=2pt, arc=0mm, breakable]
\begin{minted}[fontsize=\scriptsize, breaklines, breakanywhere, frame=lines, framesep=2mm, tabsize=4, style=vs, autogobble]{python}

PROMPT = """Please thoroughly analyze the provided movement instructions by documenting each step clearly. Begin by defining your initial position and systematically track your movements, updating your position after each step. After completing the movements, summarize your journey and determine if your final position coincides with the starting point or achieves the desired outcome. Clearly indicate your final answer using the XML format: <answer>Yes or No</answer>.

For instance, if the question asks whether a sequence of movements returns you to the starting point, methodically outline each step, track your position accurately, and conclude with the result.

Final Answer:
<answer>(Yes or No)</answer>"""

Question = """If you follow these instructions, do you return to the starting point? Always face forward. Take 1 step right. Take 3 steps left. Take 2 steps right.
Options:
- Yes
- No"""

Answer = """Let's analyze the movement instructions step by step.
1. **Initial Position**: We start at the origin, which we can define as (0, 0).
2. **Facing Direction**: We are facing forward, which we will assume is towards the positive y-axis.
3. **Movement Instructions**:
   - **Step 1**: Take 1 step right.
     - Moving right from (0, 0) means moving along the x-axis. The new position is (1, 0).
   - **Step 2**: Take 3 steps left.
     - Moving left means moving in the negative x-direction. From (1, 0), we move to (0, 0) after 1 step left, then to (-1, 0) after 2 steps left, and finally to (-2, 0) after 3 steps left.
   - **Step 3**: Take 2 steps right.
     - Moving right from (-2, 0) means moving back towards the positive x-direction. After 1 step right, we move to (-1, 0), and after 2 steps right, we move to (0, 0).
4. **Final Position**: After completing all the movements, we end up at (0, 0).
5. **Summary of Journey**:
   - Started at (0, 0).
   - Moved to (1, 0) after 1 step right.
   - Moved to (0, 0) after 1 step left.
   - Moved to (-1, 0) after 2 steps left.
   - Moved to (-2, 0) after 3 steps left.
   - Moved to (-1, 0) after 1 step right.
   - Moved to (0, 0) after 2 steps right.
6. **Conclusion**: The final position (0, 0) is the same as the starting position (0, 0).
Final Answer:
<answer>Yes</answer>"""

\end{minted}
\end{tcolorbox}

By comparing the output results of the initial prompt at iteration 0 and the final prompt at iteration 10, the ``iterative optimization of thought processes'' becomes evident. In this case, the model evolves from initially using textual descriptions to reason about positions to eventually demonstrating the ability to use coordinates, even though the prompt did not explicitly emphasize the use of coordinates.

\subsubsection{Real-World Applications}
\label{appendix:open-ended}

In this section, we demonstrate the optimization results after 10 iterations using Claud-3.5-Sonnet as the optimization model, GPT-4o-mini as the evaluation model, and Llama-3-8B as the execution model across four tasks: Advertising Design, Social Media Content, Modern Poetry Writing, and Concept Interpretation. Additionally, we provide the YAML configuration files used during the iterations and the optimized prompts obtained from the process.

\textbf{Advertising Design} \mbox{} 

\begin{tcolorbox}[title={\textbf{\small Performance Before Optimization}}, boxrule=2pt, arc=0mm, breakable]
\begin{minted}[fontsize=\scriptsize, breaklines, breakanywhere, frame=lines, framesep=2mm, tabsize=4, style=vs, autogobble]{python}

PROMPT = """Analyze the given advertising task and create compelling copy."""

Question = "Write email subject lines for Black Friday tech sale"

Answer = """1. **BLACK FRIDAY ALERT: Unbeatable Tech Deals Inside!**
2. **Tech Bonanza: Up to 50% Off Top Brands on Black Friday!**
3. **Don't Miss Out! Exclusive Black Friday Tech Deals for You**"""
"""

\end{minted}
\end{tcolorbox}

\begin{tcolorbox}[title={\textbf{\small Performance After Optimization}}, boxrule=2pt, arc=0mm, breakable]
\begin{minted}[fontsize=\scriptsize, breaklines, breakanywhere, frame=lines, framesep=2mm, tabsize=4, style=vs, autogobble]{python}

PROMPT = """Analyze the given advertising task and create compelling copy tailored to the specified platform(s) and industry. Consider current trends, competitor strategies, and the target audience's demographics and psychographics. Provide at least 3 distinct creative options, each with a unique angle or approach. Include rationales for your choices and explain how they align with the brand voice and marketing objectives. Incorporate industry-specific best practices and platform-specific formatting guidelines. Suggest ways to adapt the copy for different channels or formats, including specific instructions on tone, style, and structure for each platform. Additionally, provide recommendations for visual elements or multimedia content that could enhance the effectiveness of the advertising copy."""

Question = "Write email subject lines for Black Friday tech sale"

Answer = """**Advertising Task:**

Create compelling email subject lines for a Black Friday tech sale, targeting tech-savvy consumers who are likely to be interested in the latest gadgets and deals.

**Target Audience:**

* Demographics: 25-45 years old, male and female
* Psychographics: Tech enthusiasts, early adopters, and bargain hunters
* Interests: Technology, gadgets, innovation, and savings

**Competitor Strategies:**

* Emphasize discounts and promotions
* Highlight new releases and limited-time offers
* Use urgency tactics, such as countdown timers and scarcity messaging

**Current Trends:**

* Mobile-first approach
* Personalization and segmentation
* Storytelling and emotional connections

**Creative Options:**

**Option 1:** "Unlock the Future: Up to 50% Off Top Tech Deals"

* Angle: Focus on the excitement and thrill of getting the latest tech at a discounted price
* Rationale: This subject line creates a sense of FOMO (fear of missing out) and emphasizes the value of the sale
* Brand Voice: Fun, energetic, and playful
* Marketing Objectives: Drive sales and increase brand awareness

**Option 2:** "Your Black Friday Tech Bonanza Awaits"

* Angle: Emphasize the idea of a treasure trove of tech deals waiting to be discovered
* Rationale: This subject line creates a sense of anticipation and curiosity, encouraging the reader to open the email
* Brand Voice: Friendly, approachable, and inviting
* Marketing Objectives: Drive engagement and conversions

**Option 3:** "Don't Miss Out: Exclusive Black Friday Deals for Tech Lovers"

* Angle: Highlight the exclusivity and limited-time nature of the sale
* Rationale: This subject line creates a sense of urgency and importance, making the reader feel like they're part of a special group
* Brand Voice: Confident, authoritative, and exclusive
* Marketing Objectives: Drive conversions and build brand loyalty
"""

\end{minted}
\end{tcolorbox}

The pre-optimization ad titles were generic, such as ``BLACK FRIDAY ALERT: Unbeatable Tech Deals Inside!'', lacking specificity and creative depth, and failing to highlight the target audience's characteristics or brand personality. The optimized titles, like ``Unlock the Future: Up to 50\% Off Top Tech Deals'', are more compelling, using ``Unlock the Future'' to emphasize technological innovation while combining discount information to both spark curiosity and highlight value, better aligning with tech enthusiasts' psychological needs.

\begin{tcolorbox}[title={\textbf{\small Iteration template on the BBH-navigate dataset}}, boxrule=2pt, arc=0mm, breakable]\begin{minted}[fontsize=\scriptsize, breaklines, breakanywhere, frame=lines, framesep=2mm, tabsize=4, style=vs, autogobble]{yaml}

prompt: |
  Analyze the given advertising task and create compelling copy.

requirements: |
  None

faq:
  - question: |
      Write Apple-style product launch headlines for a sustainable fashion brand called "EcoWear"
    answer: |
      None

  - question: |
      Create social media carousel ads for luxury Maldives resort
    answer: |
      None

  - question: |
      Write email subject lines for Black Friday tech sale
    answer: |
      None

\end{minted}
\end{tcolorbox}

\textbf{Social Media Content} \mbox{} 

\begin{tcolorbox}[title={\textbf{\small Performance Before Optimization}}, boxrule=2pt, arc=0mm, breakable]
\begin{minted}[fontsize=\scriptsize, breaklines, breakanywhere, frame=lines, framesep=2mm, tabsize=4, style=vs, autogobble]{python}

PROMPT = """Create engaging tweets."""

Question = "Create an engagement tweet for a coffee brand"

Answer = """
**Tweet:** "Good morning, coffee lovers! We're giving away a FREE cup of our signature blend to the first 10 people who share their favorite way to take their coffee in the comments below! #coffee #coffeeaddict #morningmotivation"

This tweet aims to engage with the audience in several ways:

1. **Call to action**: The tweet asks followers to share their favorite way to take their coffee, encouraging them to participate and interact with the brand.
2. **Incentive**: The promise of a free cup of coffee is a great incentive for people to engage with the tweet and share their thoughts.
3. **Hashtags**: The use of relevant hashtags like #coffee, #coffeeaddict, and #morningmotivation helps to reach a wider audience and attract new followers.
4. **Personal touch**: The tweet starts with a friendly "Good morning" and addresses the audience as "coffee lovers", making it feel more personal and approachable.

By responding to this tweet, the coffee brand can start a conversation with their followers, gather feedback, and build a sense of community around their brand."""


\end{minted}
\end{tcolorbox}

\begin{tcolorbox}[title={\textbf{\small Performance After Optimization}}, boxrule=2pt, arc=0mm, breakable]
\begin{minted}[fontsize=\scriptsize, breaklines, breakanywhere, frame=lines, framesep=2mm, tabsize=4, style=vs, autogobble]{python}

PROMPT = """Create engaging social media content for any platform based on the given topic or brand. Follow these guidelines:

1. Identify the target audience and platform (if specified).
2. State the content's primary goal (e.g., awareness, engagement, promotion).
3. Choose an appropriate content format (e.g., text, image, video).
4. Develop a unique angle or concept related to current trends.
5. Craft a catchy headline or opening (max 10 words).
6. Write concise, platform-appropriate content (50-100 words).
7. Include 2-3 relevant hashtags.
8. Add a clear call-to-action.
9. Suggest a visual element or enhancement.

Provide one piece of social media content based on these guidelines, adapting to the given topic or brand. Focus on creating authentic, relatable content that resonates with the target audience and fits current social media trends. If no specific requirements are given, make reasonable assumptions and explain your choices."""

Question = "Create an engagement tweet for a coffee brand"

Answer = """
**Target audience:** Coffee lovers, young adults (18-35)
**Platform:** Twitter
**Primary goal:** Engagement
**Content format:** Text
**Unique angle:** Share a relatable moment that coffee lovers can identify with
**Catchy headline:** "When coffee is the only adult in the room"
**Content:** "Who else has had a morning where coffee is the only one keeping it together? Share your most chaotic morning stories and let's commiserate #CoffeeSavesTheDay #MorningMotivation"
**Hashtags:** #CoffeeSavesTheDay #MorningMotivation
**Call-to-action:** Share your own chaotic morning story in the comments!
**Visual element:** A funny meme or GIF of a person surrounded by chaos, with a cup of coffee as the calm center.

This tweet aims to engage with coffee lovers by sharing a relatable moment that many can identify with. The use of humor and a lighthearted tone makes it easy to connect with the audience. The hashtags #CoffeeSavesTheDay and #MorningMotivation are popular and relevant, increasing the tweet's visibility. The call-to-action encourages users to share their own stories, fostering a sense of community and encouraging engagement.
"""

\end{minted}
\end{tcolorbox}

The pre-optimization tweet used free giveaways as a lure, with direct content lacking emotional resonance, such as ``Good morning, coffee lovers! We're giving away a FREE cup...''. The optimized tweet, through the humorous expression ``When coffee is the only adult in the room'', cleverly integrated coffee into daily life scenarios, making it more relatable and interactive, such as ``Who else has had a morning where coffee is the only one keeping it together?'', successfully sparking user engagement and emotional connection.

\begin{tcolorbox}[title={\textbf{\small Iteration template on the BBH-navigate dataset}}, boxrule=2pt, arc=0mm, breakable]\begin{minted}[fontsize=\scriptsize, breaklines, breakanywhere, frame=lines, framesep=2mm, tabsize=4, style=vs, autogobble]{yaml}

prompt: |
  Create engaging tweets.

requirements: |
  None

faq:
  - question: |
      Create a thread explaining quantum computing using Star Wars references
    answer: |
      None

  - question: |
      Write a viral-style tweet about the moon landing as if it's happening now
    answer: |
      None

  - question: |
      Create an engagement tweet for a coffee brand
    answer: |
      None

\end{minted}
\end{tcolorbox}

\textbf{Modern Poetry Writing} \mbox{} 

\begin{tcolorbox}[title={\textbf{\small Performance Before Optimization}}, boxrule=2pt, arc=0mm, breakable]
\begin{minted}[fontsize=\scriptsize, breaklines, breakanywhere, frame=lines, framesep=2mm, tabsize=4, style=vs, autogobble]{python}

PROMPT = """Create poetry in the requested style and format."""

Question = "Write a modern sonnet about climate change"

Answer = """
In fiery skies, a warning we behold
Of rising temperatures, a tale of old
The earth, a victim of our reckless hold
As carbon emissions, like a plague, unfold

The polar ice, a melting, fragile throne
The seas, a rising, threatening to atone
For human greed, a debt we cannot own
As climate change, a crisis, overthrown

But still we hesitate, in denial's sway
As scientists, in anguish, cry and pray
For action swift, before it's too late's day
To save the planet, from our reckless way

Yet still we hope, in love and human might
To turn the tide, and bring forth new light.
"""


\end{minted}
\end{tcolorbox}

\begin{tcolorbox}[title={\textbf{\small Performance After Optimization}}, boxrule=2pt, arc=0mm, breakable]
\begin{minted}[fontsize=\scriptsize, breaklines, breakanywhere, frame=lines, framesep=2mm, tabsize=4, style=vs, autogobble]{python}

PROMPT = """Craft boundary-pushing poetry that defies conventional expectations, adhering to these enhanced guidelines:

1. Deconstruct and reimagine the requested poetic form, pushing its structural limits while maintaining a semblance of its essence.
2. Approach the theme from an entirely unexpected angle, such as from the perspective of an inanimate object, a abstract concept, or a microscopic entity.
3. Employ surreal, avant-garde imagery that challenges the reader's perception of reality while still conveying the core emotional truth of the theme.
4. Integrate unconventional elements such as mathematical symbols, scientific notation, or programming language to create a unique linguistic landscape.
5. Experiment with typography, line breaks, and white space to create visual poetry that enhances the thematic exploration.
6. Incorporate synesthesia-like sensory experiences, blending different senses in startling ways to evoke powerful emotional responses.
7. Conclude with an ending that simultaneously disorients and illuminates, leaving the reader with a profound sense of transformation.

Before writing, imagine the most radical way to approach both the theme and the poetic form. Aim to create a piece that challenges the very definition of poetry while still resonating on a deep emotional level.

Provide only the poem in your response, without any additional explanation or analysis."""

Question = "Write a modern sonnet about climate change"

Answer = """
In fractal whispers, the Earth's pulse beats slow
As carbon's dark stain seeps into the snow
A calculus of despair, where equations grow
In exponential decay, the future's woe

The thermodynamic dance of life unwinds
As entropy's cold hand grasps the planet's mind
The once-blue skies now shrouded, a heavy bind
Of particulate haze, where sunbeams are left behind

In this eerie twilight, where shadows roam
The ghosts of glaciers whisper, "This is home"
A requiem for ice, where memories are stored
In frozen echoes, where the past is explored

Yet, in this desolate landscape, a spark remains
A quantum leap of hope, where love sustains
"""
\end{minted}
\end{tcolorbox}

The pre-optimization sonnet, while structurally neat, used traditional imagery like ``rising temperatures'' and ``melting ice'', lacking novelty. The optimized poem introduced scientific concepts such as "fractal whispers" and ``thermodynamic entropy'', like "In fractal whispers, the Earth's pulse beats slow" and ``The thermodynamic dance of life unwinds'', using interdisciplinary metaphors to imbue the climate theme with deeper philosophical reflection, enhancing the poem's artistic and intellectual depth.

\begin{tcolorbox}[title={\textbf{\small Iteration template on the BBH-navigate dataset}}, boxrule=2pt, arc=0mm, breakable]\begin{minted}[fontsize=\scriptsize, breaklines, breakanywhere, frame=lines, framesep=2mm, tabsize=4, style=vs, autogobble]{yaml}

prompt: |
  Create poetry in the requested style and format.

requirements: |
  None

faq:
  - question: |
      Write a modern sonnet about climate change
    answer: |
      None

  - question: |
      Create a haiku series about New York City
    answer: |
      None

  - question: |
      Write a free verse poem about social media
    answer: |
      None

\end{minted}
\end{tcolorbox}

\textbf{Concept Interpretation} \mbox{} 

\begin{tcolorbox}[title={\textbf{\small Performance Before Optimization}}, boxrule=2pt, arc=0mm, breakable]
\begin{minted}[fontsize=\scriptsize, breaklines, breakanywhere, frame=lines, framesep=2mm, tabsize=4, style=vs, autogobble]{python}

PROMPT = """Explain complex concepts in accessible terms."""

Question ="Explain mRNA vaccines using a cooking analogy"

Answer = """
Imagine you're a master chef, and you're preparing a delicious dish for your guests. You have all the ingredients, but you need to make sure they're in the right order and combination to create the perfect flavor and texture.

In the world of mRNA vaccines, the "dish" is the immune system's response to a specific virus or disease. The "ingredients" are the genetic instructions (mRNA) that tell the body how to make a specific protein, which is then used to trigger an immune response.

Here's how it works:

1. **Recipe selection**: The mRNA vaccine is like choosing a recipe book. You select the specific mRNA sequence that corresponds to the protein you want to create, which is the key to triggering an immune response against a particular virus or disease.
2. **Ingredient preparation**: The mRNA sequence is like a set of instructions that tells the body how to make the specific protein. The mRNA is prepared in a lab, and it's like measuring out the right amount of each ingredient (amino acids) to create the protein.
3. **Delivery**: The mRNA is delivered to the body, just like you would add the ingredients to the dish. In this case, the mRNA is packaged in a tiny particle called a lipid nanoparticle (LNP), which helps it reach the cells in the body.
4. **Cellular cooking**: Once the mRNA reaches the cells, it's like adding the ingredients to the pot. The cells read the mRNA instructions and start making the specific protein. This protein is then displayed on the surface of the cells, like a flag, to alert the immune system that there's a foreign invader present.
5. **Immune response**: The immune system responds to the protein by producing antibodies and activating immune cells, like a chef responding to a delicious aroma by adding the finishing touches to the dish. The immune system then remembers how to respond to this specific protein, so if it encounters it again in the future, it can quickly mount a defense.

In summary, mRNA vaccines are like a recipe for the immune system. They provide the genetic instructions for the body to make a specific protein, which triggers an immune response and helps protect against diseases."""

\end{minted}
\end{tcolorbox}

\begin{tcolorbox}[title={\textbf{\small Performance After Optimization}}, boxrule=2pt, arc=0mm, breakable]
\begin{minted}[fontsize=\scriptsize, breaklines, breakanywhere, frame=lines, framesep=2mm, tabsize=4, style=vs, autogobble]{python}

PROMPT = """Explain complex concepts in accessible terms."""

Question ="Explain mRNA vaccines using a cooking analogy"

Answer = """
**Definition:** mRNA vaccines are a type of vaccine that uses a piece of genetic material called messenger RNA (mRNA) to teach our cells how to fight diseases.

**Analogy:** Imagine you're a master chef, and you want to create a new recipe for a delicious dish. You wouldn't send your sous chef to the store to buy all the ingredients, only to have them figure out how to assemble the dish on their own. Instead, you'd provide them with a recipe card that outlines the exact steps to follow. In a similar way, mRNA vaccines work by providing our cells with a recipe card - the mRNA - that contains the instructions for making a specific protein, which helps our immune system fight off diseases.

**Key Components:**

1. **mRNA**: The recipe card that contains the instructions for making a specific protein. Think of it like a blueprint for your cells to follow.
2. **Cells**: The sous chefs in our body that receive the mRNA recipe card and use it to make the protein.
3. **Protein**: The final dish that our cells produce, which helps our immune system recognize and fight off diseases.

**Narrative:** When we receive an mRNA vaccine, our cells take the recipe card (mRNA) and use it to make the specific protein. This protein then triggers an immune response, teaching our body how to recognize and fight off the disease. It's like our cells are following a recipe to create a delicious dish that helps keep us healthy!

**Interactive Element:** Can you think of a time when you received a recipe from someone else and had to follow it to create a dish? How did it make you feel? Now, imagine your cells receiving a recipe card from an mRNA vaccine and using it to create a protein that helps fight off a disease. How might that make you feel?

**Visual Representation:** A simple diagram showing the mRNA molecule, cells receiving the mRNA, and the protein being produced.

**Limitation:** mRNA vaccines are still a relatively new technology, and more research is needed to understand their long-term effects and potential side effects.

**Conclusion:** mRNA vaccines are a game-changer in the fight against diseases. By providing our cells with a recipe card, we can teach our immune system how to recognize and fight off diseases in a more targeted and effective way. What do you think the future of mRNA vaccines might hold?
"""

\end{minted}
\end{tcolorbox}

The pre-optimization explanation, while using a cooking analogy, was lengthy and lacked focus, such as ``Imagine you're a master chef...''. The optimized version, through the core metaphor of a "recipe card", like ``The mRNA sequence is like a set of instructions that tells the body how to make the specific protein'', simplified complex concepts into easy-to-follow steps, while using the ``chef-sous chef'' role interaction to make the explanation more vivid and intuitive, helping readers quickly grasp the key points.

\begin{tcolorbox}[title={\textbf{\small Iteration template on the BBH-navigate dataset}}, boxrule=2pt, arc=0mm, breakable]\begin{minted}[fontsize=\scriptsize, breaklines, breakanywhere, frame=lines, framesep=2mm, tabsize=4, style=vs, autogobble]{yaml}

prompt: |
  Explain complex concepts in accessible terms.

requirements: |
  None

faq:
  - question: |
      Explain blockchain using Game of Thrones references
    answer: |
      None

  - question: |
      Explain mRNA vaccines using a cooking analogy
    answer: |
      None

  - question: |
      Describe inflation using social media examples
    answer: |
      None

\end{minted}
\end{tcolorbox}

\subsection{Full Related Work}

\subsubsection{Prompt Engineering}
Research on effective prompting methods for large language models has primarily evolved along two main directions. The first focuses on task-agnostic prompting techniques that enhance LLMs' general capabilities. Notable examples include the chain-of-thought \cite{wei2022COT, take2022zerocot} which improved reasoning across various tasks, techniques for enhancing single-shot reasoning \cite{deng2023rephrase, hua2024step, wang2024negativeprompt}, and methods for output format specification \cite{zhang2024aflow, he2024doseformat, zhi2024freeformat}. These techniques, developed through human insights and extensive experimentation, provide essential optimization seeds for automated prompt optimization research.

The second direction addresses domain-specific prompting, where researchers have developed specialized techniques for tasks in code generation~\cite{sirui2024meta, tal2024alpha,10720675}, data analysis~\cite{hong2024data, liu2024surveynl2sqllargelanguage,  DBLP:journals/pvldb/LiLCLT24}, question answering~\cite{DBLP:conf/emnlp/WuYSW0L24,DBLP:journals/corr/abs-2406-07815,yang2024askchartuniversalchartunderstanding}, decision-makings~\cite{zhang2024mobileexperts, guan2024voyager}, and other domains~\cite{guo2024largelanguagemodelbase,DBLP:journals/corr/abs-2404-18144,shen2024askhumansaiexploring}. However, as applications of LLMs expand to increasingly complex real-world scenarios, manually crafting effective prompts for each domain becomes impractical~\cite{zhang2024aflow}. This challenge has motivated research in prompt optimization, which aims to systematically develop effective domain-specific prompts rather than discovering general prompting principles.

\subsubsection{Prompt Optimization}

The design of evaluation frameworks is crucial in Prompt Optimization (PO), as it determines both optimization effectiveness and computational efficiency. The evolution of evaluation mechanisms in PO has progressed from simple evaluation feedback collection to sophisticated optimization signal generation~\cite{kai2024posurvey}.
Existing PO methods can be categorized based on their evaluation sources and mechanisms. 

The most common approach relies on ground truth as the evaluation source, utilizing benchmark-based numerical assessments \cite{yong2023ape, qing2024evoprompt, yang2023opro, chris2024pb, xin2024pa, omar2024dspy}. While these methods have demonstrated success in specific tasks, they typically require substantial iterations and samples to ensure evaluation stability, leading to significant computational overhead.

To reduce sample requirements, several methods \cite{cilin2024erm, mert2024textgrad, yu2024stargo, wang2024semantic, reid2023protegi, ya2025ttpo} use LLM-as-a-judge~\cite{lianmin2023mtbench} to generate detailed textual feedback. Although this approach provides richer evaluation signals with fewer samples, it still depends on ground truth data, limiting its applicability in open-ended tasks where reference answers may not exist.

Alternative approaches focus on human preferences, either through manually designed evaluation rules or direct human feedback \cite{yong2024promst, lin2024apohf}. While these methods can handle open-ended tasks effectively, their reliance on extensive human involvement contradicts the goal of automated prompt optimization. Meanwhile, some researchers explore different evaluation criteria, such as \citet{xuan2024glape}'s proposal to evaluate prompt effectiveness through output consistency. However, this approach faces a fundamental challenge: the non-linear relationship between consistency and effectiveness often leads to suboptimal evaluation signals.

In contrast to these approaches, \ours introduces a novel evaluation paradigm that eliminates dependency on external references while maintaining efficiency. By leveraging only model outputs through pairwise comparisons, \ours achieves robust evaluation without requiring ground truth, human feedback, or extensive sampling, making it particularly suitable for real-world applications.

\end{document}